\documentclass[10pt,journal,compsoc]{IEEEtran}
\ifCLASSOPTIONcompsoc
  \usepackage[nocompress]{cite}
\else
  \usepackage{cite}
\fi
\ifCLASSINFOpdf
\else
\fi
\usepackage{indentfirst}
\usepackage{amssymb}
\usepackage{amsmath}
\usepackage{theorem}
\usepackage{graphicx}
\usepackage[sectionbib]{bibunits}
\usepackage{subfigure}
\usepackage{wrapfig}
\usepackage{graphics}
\usepackage{epsfig}
\usepackage{url}
\usepackage{makeidx}
\usepackage{showidx}
\usepackage{verbatim}
\usepackage[algo2e,vlined,ruled]{algorithm2e}
\usepackage{url}
\usepackage{xspace}
\usepackage{alltt}
\usepackage{showidx}
\usepackage{psfrag}
\usepackage{fancyvrb}
\usepackage{color}
\usepackage{algorithm}
\usepackage{rotating}
\usepackage{moreverb}
\usepackage{relsize}
\usepackage{multirow}
\usepackage{makecell}
\usepackage{array}
\usepackage{ragged2e}

\begin{document}
%
\title{A Survey on Spatial and Spatiotemporal Prediction Methods}
%
%
%
%

\author{Zhe~Jiang
\IEEEcompsocitemizethanks{\IEEEcompsocthanksitem Z. Jiang was with the Department of Computer Science at the University of Alabama, Tuscaloosa,
AL, 35487. 
Email: zjiang@cs.ua.edu
}
}

\IEEEtitleabstractindextext{%
\begin{abstract}
With the advancement of GPS and remote sensing technologies, large amounts of geospatial and spatiotemporal data are being collected from various domains, driving the need for effective and efficient prediction methods. Given spatial data samples with explanatory features and targeted responses (categorical or continuous) at a set of locations, the problem aims to learn a model that can predict the response variable based on explanatory features. The problem is important with broad applications in earth science, urban informatics, geosocial media analytics and public health, but is challenging due to the unique characteristics of spatiotemporal data, including spatial and temporal autocorrelation, spatial heterogeneity, temporal non-stationarity, limited ground truth, and multiple scales and resolutions. This paper provides a systematic review on principles and methods in spatial and spatiotemporal prediction. We provide a taxonomy of methods categorized by the key challenge they address. For each method, we introduce its underlying assumption, theoretical foundation, and discuss its advantages and disadvantages. Our goal is to help interdisciplinary domain scientists choose techniques to solve their problems, and more importantly, to help data mining researchers to understand the main principles and methods in spatial and spatiotemporal prediction and identify future research opportunities.

\end{abstract}

\begin{IEEEkeywords}
Spatial and spatiotemporal prediction, survey, classification and regression, spatial big data, deep learning
\end{IEEEkeywords}}

\maketitle

\IEEEdisplaynontitleabstractindextext

%
\IEEEpeerreviewmaketitle

\IEEEraisesectionheading{\section{Introduction}\label{sec:introduction}}

%
%
%
%


\IEEEPARstart{G}{iven} spatial data samples with explanatory features and a targeted response variable at different locations and time, the spatial and spatiotemporal prediction problem aims to learn a model that can predict the response variable based on their explanatory features. Examples of the problem existed even before the popularity of computers. The spatial interpolation method Kriging, named after spatial statistician and mining engineer Krige~\cite{krige1951statistical}, was developed in the 1960s to estimate minerals for mining activities. However, with the recent advancement of GPS and remote sensing technology, as well as the popularity of geographic information system and mobile devices, large amounts of spatial and spaiotemporal data are being collected at an increasing speed, such as earth observation imagery, gauge observations in river streams, and geo-social media data. Utilizing the rich geospatial data plays a critical role in addressing many grand societal issues, but is also technically challenging due to the unique characteristics of spatial data. There is an urgent need for effective and efficient prediction methods that can unlock the values of such rich geospatial data assets.

Over the years, various spatial and spatiotemporal prediction methods have been developed by different research communities, including spatial statistics, spatial econometrics, data mining and machine learning, computer vision, remote sensing, geographic information science, and spatial database. Particularly, progress has been made with the rapid development of spatial data mining, a field that studies how to automatically identify non-trivial, previously unknown but potentially useful patterns from large geospatial datasets~\cite{shekhar2011identifying}. Yet methods developed by different research communities tend to use different vocabularies and solve problems in different angles. A systematic review that compares these methods is missing.

To this end, this paper provides a systematic review of different spatial prediction methods. We categorize methods based on the unique challenges they address, discuss their underlying assumptions, and compare their advantages and disadvantages. The goal is to help interdisciplinary researchers choose appropriate techniques to solve problems in their applications domains, and more importantly, to help data mining researchers understand the basic principles as well as identify open research opportunities in spatial prediction.

\subsection{Societal Applications}
Spatial prediction is of great importance in societal applications related to various agencies, such as the National Aeronautics and Space Administration (NASA), the National Oceanic and Atmospheric Administration (NOAA), the Department of Defense, the Department of Transportation, the Department of Homeland Security, the United States Department of Agriculture (USDA), and the Environmental Protection Agency (EPA). 
Here we categorize application examples into four major areas including earth science, urban informatics, geosocial media analytics, and public health.

\emph{Earth science}: Earth science is a major application area for spatial prediction~\cite{jiang2017spatial}. Remote sensors from satellites, airplanes, and unmanned aerial vehicles (UAVs) have collected petabytes of geo-referenced earth imagery. Particularly, the recent deployment of CubeSat fleets by commercial companies (e.g., Planet Labs Inc.) help collect high-resolution imagery that covers the entire earth surface almost every day. In addition, numerous ground sensors deployed on land or rivers collect real-time information on soil properties, river flow volume, and air quality. 
Spatial prediction on earth data plays an important role in mapping land use and land cover, understanding global deforestation~\cite{hansen2013high}, monitoring surface water dynamics~\cite{pekel2016high}, improving the situational awareness during disasters (e.g., mapping hurricane flood and earthquakes)~\cite{brivio2002integration,jiangkdd2018}, predicting crop yield~\cite{moran1997opportunities}, estimating the spatial distribution of species~\cite{austin2002spatial,elith2009species}, and mapping soil properties~\cite{chang2001near,hengl2004generic}.

\emph{Urban informatics}: Another important application area is urban informatics. Relevant spatial data includes temporally detailed road networks with real-time travel costs on individual road segments, GPS trajectories of taxis and trucks, data from video cameras and high-resolution sensors on traffic volume and occupancy close to highway intersections, passenger transactions on public transit systems such as subways and buses, high-resolution street view imagery, geo-referenced crime and accident records, and cellphone location history collected from communication towers. 
Spatial prediction plays an important role in routing and navigation services (e.g., speed profile and travel volume prediction)~\cite{Meng2017}, spatially detailed demand forecasting for sharing economy (e.g., ride-hailing, bike sharing)~\cite{demandPredAAAI18}, law enforcement management (e.g., patrol route planning targeted at predicted crime or crash hotspots), monitoring environment pollution, as well as predictive maintenance of critical urban infrastructures.

\emph{Geosocial media}: Geosocial media analytics is an important emerging application area. With the popularity of smart phones and mobile apps, large amounts of geo-referenced social media data are collected from billions of users. Examples include geo-tagged tweets and Facebook posts, geo-tagged photos and videos, online articles with named entities for locations, as well as check-in records. Geosocial media 
provides a new way to collect near real-time information about what is happening on the earth surface at a large scale and with low costs. Spatial prediction on geosocial media data has been applied to real-time spatial event detection (e.g., earthquake, flood, landslide) for disaster management~\cite{sakaki2010earthquake,zhang2017triovecevent}, spatiotemporal event forecasting (e.g., political unrest)~\cite{DBLP:conf/kdd/ZhaoSYCLR15}, and travel destination recommendations in tourism~\cite{majid2013context}. With the area growing rapidly, more applications are being developed in agriculture, environment monitoring, transportation, education, and finance. 

\emph{Public health}: Public health has long been an important application area for spatial prediction. Examples of spatial data related to public health includes demographic information on district blocks, electronic health records with patient home addresses, aggregated disease count maps at city, county or state level, population mobility data, environmental data such as air quality and water quality, and other online data such as search engine queries related to diseases. Spatial prediction plays a critical role in automatic medical diagnosis from MRI imagery, monitoring infectious disease and mapping disease risk~\cite{best2005comparison}, detecting disease outbreak, analyzing environmental factors that cause diseases~\cite{rappaport2010environment}, understanding drug epidemics~\cite{lee2016mind}, as well as providing early alert for individuals on environmental triggers of asthma.

It is important to note that the application areas listed above are not isolated but cross-cutting with each other. For example, spatial prediction on geosocial media data can help mapping flood disasters in earth science applications, detecting damage and failures of critical urban infrastructures, and monitoring disease transmission in public health. Earth observation data has been used in land use modeling for urban planning, and in modeling environment factors for public health. Such cross-cutting applications often represent new interdisciplinary research opportunities.

\subsection{Input Spatial Data}

\emph{Spatial data representation}: A geographic information system (GIS)~\cite{worboys2004gis} represents spatial data in two ways: \emph{object} and \emph{field}. In the object representation, spatial data consists of identifiable geometric objects including points, lines, and polygons. For example, cities are often represented as points on a map, while rivers and states are represented as line-strings and polygons respectively. In the field representation, spatial data consists of a spatial framework that tessellates continuous space into regular or irregular cells, together with a function that maps each cell into a value. Examples include earth observation imagery and a county-level median house income map.

\begin{table}
    \centering
    \caption{Math symbols and descriptions}
    \label{tab:symbols}
    \begin{tabular}{cp{0.72in}p{1.8in}}\hline
        Symbol & Domain & Description \\ \hline
        $n$ & $\mathbb{N}$ & The number of sample locations\\ \hline
        $\mathbf{s}$ & $\mathbb{R}^{2\times1}$ & Sample location coordinates\\ \hline
        $\mathbf{x}(\mathbf{s})$ & $\mathbb{R}^{m\times1}$ & $m$-dimensional feature vector of a sample at location $\mathbf{s}$\\ \hline
        $y(\mathbf{s})$ & $\mathbb{R}$ or $\mathcal{C}$ & Continuous or categorical response of a sample at location $\mathbf{s}$\\ \hline
        $\widehat{y}(\mathbf{s})$ & $\mathbb{R}$ or $\mathcal{C}$ & Predicted response of sample at location $\mathbf{s}$, $\mathcal{C}$ is the set of class categories\\ \hline        
        $\mathbf{X}$ &$\mathbb{R}^{n\times m}$& Feature matrix of $n$ samples\\ \hline
        $\mathbf{Y}$ &$\mathbb{R}^{n\times 1}$ or $\mathcal{C}^{n\times 1}$& Response vector of $n$ samples\\ \hline
        $\widehat{\mathbf{Y}}$ &$\mathbb{R}^{n\times 1}$ or $\mathcal{C}^{n\times 1}$& Predicted response vector\\ \hline
        $\mathbf{W}$ &$\mathbb{R}_{0+}^{n\times n}$&W-matrix\\ \hline
        $W_{ij}$ &$\mathbb{R}_{0+}$& An element of W-matrix\\ \hline
        $C(\mathbf{h})$ &$\mathbb{R}^{2\times1}\mapsto\mathbb{R}_{0+}$& Covariogram function\\ \hline
        $\rho,\lambda$ &$\mathbb{R}_{0+}$ & Weight of spatial effect\\ \hline
        $\boldsymbol{\mu}$ &$\mathbb{R}^{m\times1}$ & Mean feature vector\\ \hline
        $\boldsymbol{\Sigma}$ &$\mathbb{R}^{m\times m}$ & Covariance matrix of features\\ \hline
        $\boldsymbol{\beta}$ &$\mathbb{R}^{m\times1}$ & Coefficient vector\\ \hline
        $\epsilon(\mathbf{s_i})$ &$\mathbb{R}$& Noise (residual error) of a sample\\ \hline
        $\boldsymbol{\epsilon}$ &$\mathbb{R}^{n\times 1}$ &Noise (residual error) of $n$ samples \\ \hline
        $\boldsymbol{\theta}(\mathbf{s_i})$ &$\mathbb{R}^{m\times1}$ &Coefficients for samples at $\mathbf{s_i}$ \\ \hline
        $\boldsymbol{\Theta}$ &$\mathbb{R}^{m\times n}$ & All coefficients at different locations \\ \hline
        $\boldsymbol{\Phi}$ & Set&A set of all model parameters \\ \hline
        $w(\mathbf{s_i},\mathbf{s_0})$ &$\mathbb{R}_{0+}$ & Spatial kernel weight between location $\mathbf{s_i}$ and location $\mathbf{s_0}$\\ \hline
        $\mathbf{x}(\mathbf{s},t)$ & $\mathbb{R}^{m\times1}$ & $m$-dimensional feature vector of a sample at location $\mathbf{s}$ and time $t$\\ \hline
        $y(\mathbf{s},t)$ & $\mathbb{R}$ or $\mathcal{C}$  & Continuous or categorical response of sample at location $\mathbf{s}$ and time $t$\\ \hline
        $\mathbf{X}(t)$ &$\mathbb{R}^{n\times m}$&Feature matrix of samples at time $t$ \\ \hline
        $\mathbf{Y}(t)$ &$\mathbb{R}^{n\times 1}$ or $\mathcal{C}^{n\times 1}$&Responses (observations) at time $t$  \\ \hline        
        $\mathbf{Z}(t)$ &$\mathbb{R}^{n\times 1}$ or $\mathcal{C}^{n\times 1}$&Hidden variable vector at time $t$   \\ \hline
        $\boldsymbol{e}(t)$ &$\mathbb{R}^{n\times 1}$& Residual errors for hidden process variables at time $t$ \\ \hline
        $\boldsymbol{\epsilon}(t)$ &$\mathbb{R}^{n\times 1}$& Residual errors for observation variables at time $t$ \\ \hline
        $C(\mathbf{h},r)$ &$\mathbb{R}^{2\times1}\times \mathbb{R}\mapsto\mathbb{R}_{0+}$&Spatiotemporal covariogram\\ \hline
    \end{tabular}
\end{table}

\emph{Spatial data sample}: In traditional prediction problems, input data is often viewed as a collection of sample records. Similarly, in spatial prediction, input spatial data can be viewed as a collection of spatial data samples, whereby each sample corresponds to a spatial object (e.g., point, line or polygon), or a raster cell. 
A spatial data sample has multiple non-spatial attributes, one of which is the target response variable to be predicted and the others are explanatory features. Additionally, a spatial data sample also has location information and spatial attributes (e.g., distance to another point, length of a line, area of a polygon). These additional information  distinguish spatial data samples out from traditional data samples in two important ways: first, the location information and corresponding spatial attributes can provide additional contextual features in the explanatory feature list; second and more important, implicit spatial relationships exist based on sample locations, making samples not independent and identically distributed (non-i.i.d.) For example, in ground sensor observations on soil properties, a sample corresponds to information from one geo-located sensor. Explanatory features can include soil texture, nutrient level, and moisture. The response variable can be whether a type of plant can grow at the location. 

Formally, spatial data is a set of spatial data samples $\{(\mathbf{x(s_i)}, y(\mathbf{s_i}))|i\in \mathbb{N}, 1\leq i \leq n\}$, where $n$ is the total number of samples, $\mathbf{s_i}$ is a $2$ by $1$ spatial coordinate vector for the $i$th sample, $\mathbf{x(s_i)}$ is a $m$ by $1$ explanatory feature vector ($m$ is the feature dimension), and $y(\mathbf{s_i})$ is a scalar response (it is categorical for classification, and continuous for regression). The set of spatial samples can also be written in the matrix format, $(\mathbf{X},\mathbf{Y})$, where $\mathbf{X}=[\mathbf{x(s_1)},\mathbf{x(s_2),...,\mathbf{x(s_n)}}]^T$ is a $n$ by $m$ feature matrix, and $\mathbf{Y}=[y(\mathbf{s_1}),y(\mathbf{s_2}),...,y(\mathbf{s_n})]^T$ is a $n$ by $1$ response vector.


\subsection{Formal Problem Definition}

Given a set of spatial data samples with explanatory features $\mathbf{X}=[\mathbf{x(s_1)},\mathbf{x(s_2),...,\mathbf{x(s_n)}}]^T$ and responses $\mathbf{Y}=[y(\mathbf{s_1}),y(\mathbf{s_2}),...,y(\mathbf{s_n})]^T$, the spatial prediction problem aims to learn a model (or function) $f$ such that $\mathbf{Y}=f(\mathbf{X})$. Once the model is learned, it can be used to predict the responses at other locations based on their explanatory features. The problem can be further categorized into \emph{spatial classification} for categorical response and \emph{spatial regression} for continuous response. 

For example, in earth imagery classification for land cover mapping, input spatial data samples are training pixels whose spectral band values (e.g., red, green, blue, and near-infrared bands) are explanatory features, and whose land cover classes (e.g., forest, water) are response. The output is a classification model that can predict the land cover classes of other pixels based on spectral band values.


Spatial prediction is unique from traditional prediction in data mining. In traditional prediction problems, samples are commonly assumed to be independent and identically distributed (i.i.d.). Thus, a same model can be used to predict every sample independently, i.e., $y(\mathbf{s})=f(\mathbf{x(s)})$ for $\forall \mathbf{s}$. However, the i.i.d. assumption is often violated in spatial data due to implicit spatial relationships between sample locations. According to the first law of geography, ``everything is related to everything else, but near things are more related than distant things"~\cite{tobler1970computer}. Ignoring spatial relationships can results in incorrect model assumption and poor prediction performance.

\subsection{Challenges}
Spatial prediction poses unique challenges as compared to traditional prediction due to the special characteristics of spatial data.  Here we describe these special characteristics in an intuitive way. More mathematically rigorous discussions are in Section~\ref{sec:stat}.

\emph{Spatial autocorrelation (dependency)}: According to Tobler's first law of geography~\cite{tobler1970computer}, ``everything is related to everything else, but near things are more related than distant things." In real world spatial data, nearby locations tend to resemble each other, instead of being statistically independent. For example, the temperatures of two nearby cities are often very close. This phenomenon is also called the spatial autocorrelation effect. The existence of spatial autocorrelation is a challenge because it violates a common assumption by many traditional prediction models, i.e., learning samples are independent and identically distributed (i.i.d.). Ignoring the spatial autocorrelation effect may produce prediction models that are inaccurate or inconsistent with data. For instance, when applying a linear regression model to spatial data, the residual errors are often correlated, instead of being i.i.d. as the model assumes. In earth imagery classification, running classification models that rely on the i.i.d. assumption (e.g., decision tree, random forest) often produces results with artifacts (e.g., salt-and-pepper noise)~\cite{jiang2015focal}. 

\emph{Spatial heterogeneity}: Another unique characteristic of spatial data is that sample distribution is often not identical in the entire study area, often called the spatial heterogeneity effect. Specifically, spatial heterogeneity can be reflected in two ways, including spatial non-stationarity and spatial anisotropy. Spatial non-stationarity means that sample distribution varies across different sub-regions. For example, the same spectral signature in earth image pixels may correspond to different land cover types from tropical to temperate regions. This is a challenge because a model learned from global samples may not perform well in each sub-regions (also referred to as ``ecological fallacy"~\cite{ess2001culture}). Spatial anisotropy means that spatial dependency between sample locations is non-uniform along different directions. For example, climate data distribution is often influenced by geographical terrains (e.g., mountain range), showing unique patterns along different directions. Modeling anisotropic spatial dependency is a challenge because such dependency cannot be simply modeled as a function of distance (should be a function of direction as well). 

\emph{Limited ground truth}:  Real world spatial data often contains a large amount of information on explanatory features due to advanced data collection techniques (e.g., GPS, remote sensor). However, availability of ground truth data (e.g., land cover classes) is often very limited because collecting ground truth involves sending a field crew or hiring well-trained visual interpreters, which is both expensive and time consuming. While limited ground truth is a common challenge in many other non-spatial prediction problems, the cost of collecting ground truth for spatial data is unique in that it includes not only the labeling time cost but also the time cost for the field crew to travel on the ground between sample locations. In addition, the selection of sample locations should also consider the geographical representativeness of samples for rigorous evaluations~\cite{congalton1991review}.

\emph{Multiple scales and resolutions}: The last major challenge is that spatial data often exists in multiple spatial scales or resolutions. For example, resolutions of earth observation imagery pixels range from sub-meter (high-resolution aerial photos) to over hundreds of meters (MODIS satellite image). In precision agriculture, soil properties are recorded by ground sensors at isolated point locations, spectral signatures of crops are measured in aerial imagery that covers the entire study area, and crop yields are often measured at per plot level. This poses a challenge since traditional prediction methods often assume that data samples are at the same scale or resolution. Thus, these models cannot be directly applied. A simple approach of preprocessing to aggregate data into the same scale or resolution may cause the loss of critical information. Other approaches (e.g., statistical downscaling~\cite{wilby2004guidelines}) have been studied, but mostly for particular applications such as climate science. The challenge is still largely underexplored for broad spatial prediction applications.

\subsection{Comparison with Existing Surveys}
Most existing related surveys focus on general spatial data mining. Ester et al.~\cite{ester1997spatial} and Koperski et al.~\cite{koperski1996spatial} provide early surveys on spatial data mining from a database perspective. Miller et al.~\cite{miller2009geographic} have a book on geographic data mining and knowledge discovery that contains spatial prediction as a chapter. The chapter compares a couple of common methods in case studies but does not provide a systematic survey. Shekhar et al.~\cite{shekhar2003trends,shekhar2011identifying,shekhar2015spatiotemporal} and Atluri et al.~\cite{atluri2017spatio} provide surveys on general spatial and spatiotemporal data mining, highlighting the unique challenges of mining spatial data and spatial statistical foundation but without systematic review on prediction methods. To the best of our knowledge, there is no systematic survey on spatial prediction methods in the literature. 
 
To fill in the gap, we provide a systematic review on the principles and methods on spatial prediction. We provide a taxonomy of methods based on the unique challenge they address, including spatial autocorrelation (or dependency), spatial heterogeneity, limited ground truth, and multiple spatial scales and resolutions. When introducing each method, we start with the intuition and underlying assumption, then introduce key ideas and theoretical foundation, and finally discuss its advantages and disadvantages. 
We also introduce several spatiotemporal extensions of methods. Future research opportunities are also identified.

\subsection{Scope and Outline}
Due to space limit, we only focus on spatial prediction problems in which samples have fixed locations. Prediction for moving object data such as location prediction and recommendation are beyond our scope. We also do not include methods in computer vision unless input images are geo-referenced such as earth observation imagery. We do not particularly distinguish spatial classification and regression since methods are often exchangeable through logistic transformation.

The outline of the paper is as follows.  Section~\ref{sec:stat} introduces spatial statistics foundations. Section~\ref{sec:approach} provides a taxonomy of spatial prediction methods based on the key challenge they address, and also introduces spatiotemporal extensions of methods. Future research opportunities are discussed in Section~\ref{sec:future}. Section~\ref{sec:con} concludes the paper.

\section{Spatial Statistical Foundations}\label{sec:stat}
Spatial statistics~\cite{schabenberger2005statistical,banerjee2014hierarchical,cressie2015statistics} provides a theoretical framework to do exploratory analysis and make inference on spatial data. In spatial statistics, samples corresponding to fixed point locations in continuous space are called  \emph{point reference data}, while samples corresponding to fixed cells in the field representation are called \emph{areal data}, as summarized in Table~\ref{tab:sptype}. Samples corresponding to random point locations are called \emph{spatial point process}, which are beyond our scope. This section reviews some important spatial statistics concepts that are the foundation of many spatial prediction methods, including spatial autocorrelation, stationarity, isotropy, variogram, covariogram, and spatial heterogeneity.

\begin{table}[h]
    \centering
    \caption{Types of spatial data}
    \label{tab:sptype}
    \begin{tabular}{ccc}\hline
        \multicolumn{2}{c}{Data Representation View} & Spatial Statistics View\\ \hline
        \multirow{4}{*}{Object} & \multirow{2}{*}{Points} & Point reference data\\ 
        & & Spatial point process\\ 
        & Lines & \\ 
        & Polygons & \\ \hline
        \multirow{2}{*}{Field} & Regular cells &  \multirow{2}{*}{Areal data}\\ 
         &Irregular cells & \\ \hline
    \end{tabular}
\end{table}

\subsection{Spatial Autocorrelation (Dependency)}\label{subsec:ssa}
The first law of geography indicates that spatial data samples are not statistically independent but correlated, particularly across nearby locations. This effect is also called the \emph{spatial autocorrelation} effect. We exchange the usage of ``autocorrelation" and ``dependency" on spatial data to mean the same effect. The specific statistics of spatial autocorrelation vary between areal data and point reference data, which are introduced separately below. 

\subsubsection{Spatial autocorrelation on areal data}

\emph{Spatial neighborhood and W-matrix}: On areal data, spatial data samples are regular or irregular cells in a discrete tessellation of continuous space. The range of spatial dependency is assumed to be within \emph{spatial neighborhood}, which can be defined based on cell adjacency or distance. Most often, two samples (cells) are spatial neighbors if they share boundaries (rook neighborhood), or if they share corners or boundaries (queen neighborhood). Spatial neighborhood relationships across all samples (cells) can be represented by a $n$ by $n$ square matrix called W-matrix $\mathbf{W}$, where $n$ is the number of samples, element $W_{ij}>0$ if the $i$th sample and the $j$th sample are spatial neighbors, and $W_{ij}=0$ otherwise (by default, $W_{ii}=0$, i.e., samples are not neighbors of themselves). 

\emph{Spatial autocorrelation}:
Based on the definition above, spatial autocorrelation statistics is defined as the correlation between observations of the same variable at neighboring cells. One example for continuous response 
$y$ is Moran's $I$, which is defined as
\begin{equation}
I=\frac{\sum_{i=1}^n\sum_{j=1}^n W_{ij}(y(\mathbf{s_i})-\bar{y})(y(\mathbf{s_j})-\bar{y})}
{(\sum_{i=1}^n\sum_{j=1}^n W_{ij})\sum_{i=1}^n(y(\mathbf{s_i})-\bar{y})^2/n}    
\end{equation}
where $\bar{y}=\sum_{i=1}^n y(\mathbf{s_i})/n$ with $n$ as the total number of samples (cells). Similar to Pearson's correlation, the value of Moran's $I$ is within $[-1,1]$. A positive Moran's $I$ indicates that nearby locations tend to have similar values, while a negative $I$ indicates that nearby locations tend to have different values. There are also other spatial autocorrelation statistics, such as Geary's $C$, $G$ statistic, Black-Black joint count~\cite{schabenberger2005statistical}.

\subsubsection{Spatial autocorrelation on point reference data}\label{subsubsec:sa}
Spatial autocorrelation statistics on point reference data are different from those on areal data in that they measure correlation between variables at any two locations in continuous space, only based on observations at a limited number of point locations. In order to make such measures possible, further assumptions on data distribution have to be made, including spatial stationarity and isotropy. 

\emph{Spatial stationarity}: Spatial stationarity is an assumption that sample statistical properties are location invariant. There are different levels of stationarity assumption according to which statistical properties stay invariant when locations are shifted. The strongest assumption is \emph{strict stationarity}, meaning that the joint distribution of variables at several locations stay unchanged if their locations are shifted by a same distance and direction, i.e.,
$P(y(\mathbf{s_1}),...,y(\mathbf{s_n}))\equiv P(y(\mathbf{s_1}+\mathbf{h}),...,y(\mathbf{s_n}+\mathbf{h})), \forall \mathbf{s_1}, \mathbf{s_2}, ..., \mathbf{s_n},\mathbf{h}$.
This assumption is the strongest because the joint distribution determines all other statistical properties. \emph{Weak stationarity} assumes that the first and second moments of spatial variables are invariant with location shifting, i.e., 
$E(y(\mathbf{s}))\equiv \boldsymbol{\mu}$ and $Cov(y(\mathbf{s}+\mathbf{h}),y(\mathbf{s}))\equiv C(\mathbf{h})$ for $\forall \mathbf{s},\mathbf{h}$. With the assumption of weak stationarity, the covariance between any two locations is a function $C(\mathbf{h})$ on the relative location difference $\mathbf{h}$, which is called \emph{covariogram}. Another weaker assumption is \emph{intrinsic stationarity}, formally,
$E(y(\mathbf{s}+\mathbf{h})-y(\mathbf{s}))^2\equiv \gamma(\mathbf{h})$ for $\forall\mathbf{s},\mathbf{h}$. The function $\gamma(\mathbf{h})$ here is also called \emph{variogram}. In practice, a spatial variable usually has a non-constant mean  (also called trend) $E(y(\mathbf{s}))$, so weak (or intrinsic) stationarity is often assumed on residual errors $y(\mathbf{s})-E(y(\mathbf{s}))$.

\emph{Spatial isotropy}: Weak or intrinsic stationarity assumes that the second order statistical property of a variable at two locations is a function of location difference vector only.  \emph{Spatial isotropy} further assumes that the properties only depend on distance regardless of direction, i.e., covariogram $C(\mathbf{h})\equiv C(h)$ and variogram $\gamma(\mathbf{h})\equiv \gamma(h)$, where $h=\|\mathbf{h}\|_2$. In other words, covariance between variables at any two locations only depends on their relative distance. 

In a point reference data, assuming weak stationarity and isotropy, we can empirically estimate the variogram or covariogram function by fitting a curve (e.g., linear, spherical, exponential) based on observations at several point locations. Once the curve is fitted, we can get covariance between variables at any two locations simply based on their distance.

From the discussions above, we can observe that the spatial autocorrelation effect (dependency) is measured differently on areal data and point reference data. On areal data, it is based on the definition of spatial neighborhood, while on point reference data, it is based on covariance function (covariogram). 
The existence of spatial autocorrelation poses a challenge in spatial prediction, since the common assumption that samples are statistically independent in many traditional prediction models is no longer valid. Ignoring this challenge can lead to poor prediction performance.

\subsection{Spatial heterogeneity}
If a spatial distribution is stationary and isotropic, it is called \emph{homogeneous}~\cite{banerjee2014hierarchical}. Thus, a spatial distribution is \emph{heterogeneous} if it is either non-stationarity or anisotropy. The terms of \emph{homogeneity} and \emph{heterogeneity} should not be confused with homoscedasticity and heteroscedasticity, which specifically refer to properties on variance.

From discussions in Section~\ref{subsec:ssa}, we can see that assuming spatial homogeneity (stationarity and isotropy) can greatly simplify the statistical modeling of spatial data. Thus, this assumption is made in many spatial prediction methods. However, the assumption can be violated by real world spatial data, which is often spatially heterogeneous (spatially non-stationary or anisotropic). For example, spectral features of forest, land and water in earth imagery vary from tropical regions to temperate regions. As another instance of example, when classifying earth observation imagery pixels into water and land, spatial dependency across nearby water locations is anisotropic following geographic terrain and topography (water flows from a higher elevation to a nearby lower elevation due to gravity). Spatial heterogeneity poses a challenge in spatial prediction since a model learned from an entire study area may perform poorly in some local regions.

\section{A Taxonomy of Spatial Prediction Methods}\label{sec:approach}

\begin{figure*}
    \centering
    \includegraphics[width=6in]{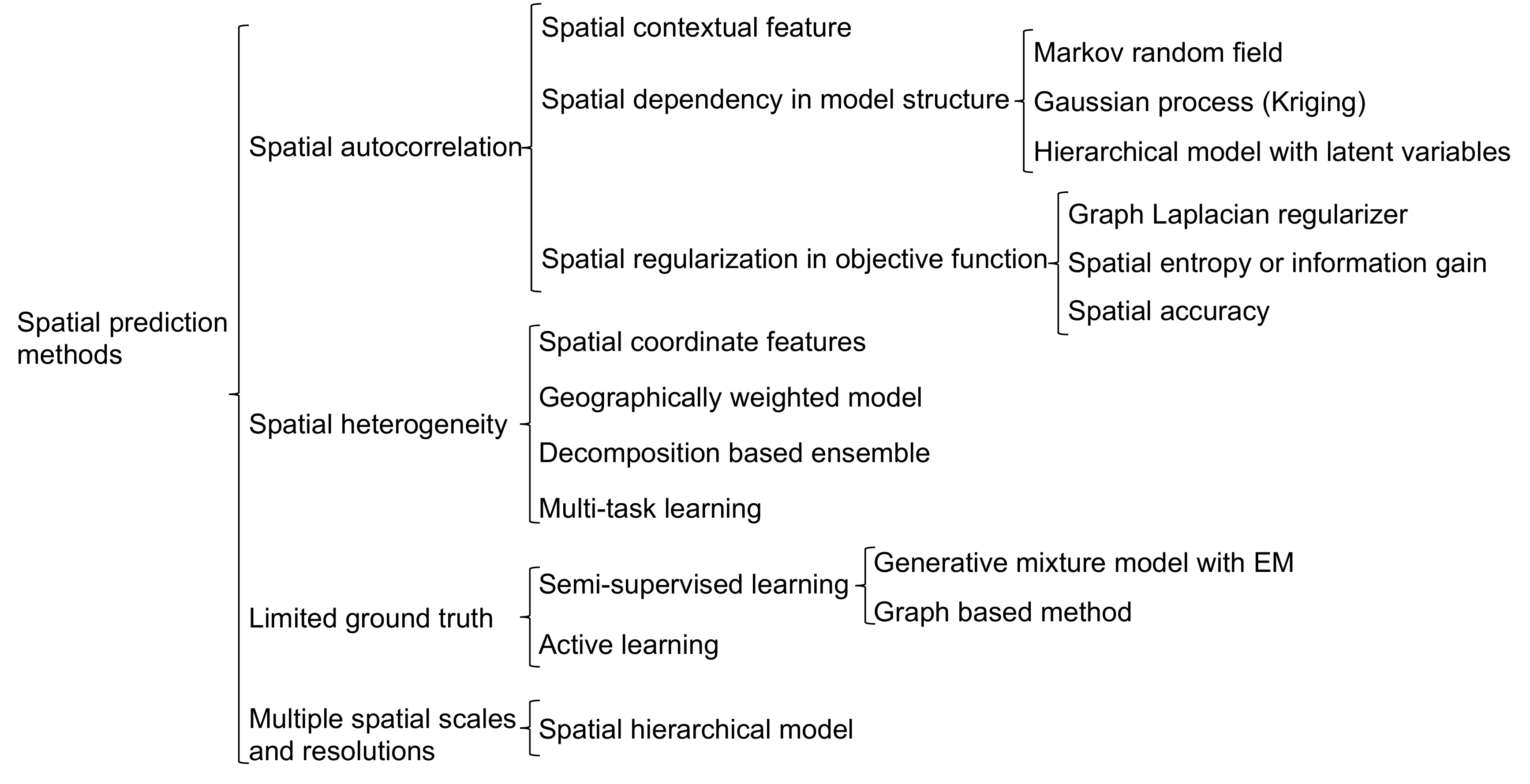}
    \caption{A taxonomy of spatial prediction methods}
    \label{fig:taxonomy}
\end{figure*}

This section provides a taxonomy of existing spatial prediction methods, as shown in Figure~\ref{fig:taxonomy}. Methods are first categorized by the unique challenge they address, including spatial autocorrelation, spatial heterogeneity, limited ground truth, and multiple spatial scales and resolutions. Within each category, we further group the methods based on the strategies they use to address the challenge. When reviewing a method, we start from the intuition behind it, highlight underlying assumptions, explain the key ideas, and discuss its advantages and disadvantages. We also discuss spatiotemporal extensions of methods in the end.

\subsection{Spatial Autocorrelation (Dependency)}\label{subsec:sa}
Addressing the challenge of spatial autocorrelation or dependency requires spatial prediction algorithms to go beyond the independence assumption. Common strategies
include spatial contextual feature generation for model inputs, spatial dependency constraint within model structure, and spatial regularization for model objective function.

\subsubsection{Spatial contextual feature generation}\label{subsubsec:scf} 
One way of incorporating spatial dependency into prediction methods is to augment input data with additional spatial contextual features. The spatial context of a sample location refers to information surrounding it, such as relationships to other objects or locations, attributes of nearby samples, auxiliary semantic information from additional data sources. Once spatial contextual information is added into explanatory features, traditional non-spatial prediction methods can be used. We now introduce several approaches to generate spatial contextual features.

\emph{Spatial relationship features}: Spatial contextual features can be generated based on spatial relationships with other locations or objects, such as distance or direction, touching, lying within or overlapping with another object. Spatial relationship features can be readily used in rule-based or decision tree-based models. Examples of techniques include spatiotemporal probability tree model to classify meteorological data on storms~\cite{mcgovern_spatiotemporal_2008,mcgovern_enhanced_2013}, multi-relational spatial classification~\cite{frank_multi-relational_2009}, prediction based on spatial association rules~\cite{ding_discovery_2009}.

\emph{Spatial contextual features on raster data}: Spatial contextual features have long been used in classifying raster data (e.g., earth observation imagery) to reduce salt-and-pepper noise~\cite{lu2007survey}. Specific methods include neighborhood window filters (e.g., median filter~\cite{chan2005salt}, weighted median filter~\cite{brownrigg1984weighted}, adaptive median filter~\cite{hwang1995adaptive}, decision-based filter~\cite{chan2005salt,esakkirajan2011removal}, etc.), spatial contextual variables and textures~\cite{puissant2005utility}, neighborhood spatial autocorrelation statistics~\cite{jiang2015focal,jiang_focal-test-based_2013}, morphological profiling~\cite{benediktsson2005classification}, and object-based image analysis (e.g., mean, variance, texture of object segments)~\cite{hay2008geographic}. Sometimes, these methods are used in the post-processing step~\cite{tarabalka2009spectral}.

\emph{Spatial contextual features from multi-source data fusion}: One unique property of spatial data is that information from different sources can be fused into the same spatial framework, providing important spatial contexts for learning samples. For example, when predicting a fine-grained air quality map for an entire city, we can generate contextual features by fusing air quality records at ground stations, weather information, road network and traffic data, as well as POIs~\cite{zheng2013u}. When predicting human behaviors from location history, auxiliary data from geosocial media can provide important semantic annotations~\cite{wu2015semantic,wu2016did}. Generating contextual features through data fusion can have its own challenges (e.g., multi-modality, sparsity, noise). Various techniques have been explored such as coupled matrix factorization, and context-aware tensor decomposition with manifold~\cite{zheng2015methodologies}.

Spatial contextual feature generation is important in many practical applications (e.g., urban computing) due to two main advantages. First, generating appropriate contextual features can significantly enhance prediction accuracy due to the effectiveness of those features in explaining the response variable. Second, after spatial contextual features are generated, many traditional non-spatial predictive models can be used (e.g., random forest, support vector machine). This is sometimes convenient since there is no need to modify non-spatial prediction models or learning algorithms. At the same time, spatial contextual feature generation may require significant knowledge about the application domain.


\subsubsection{Spatial dependency within model structure}\label{sec:sainm}
Instead of generating spatial contextual features and utilizing traditional non-spatial prediction methods, we can directly incorporate spatial dependency in model structure. There are three different strategies to do that, including Markov random field based models for areal data, Gaussian process based models for point reference data, and hierarchical models with latent variables which provide a new perspective of capturing spatial dependency for both areal data and point reference data.

\emph{(1) Markov Random Field Based Models}

Markov random field (MRF) is a widely used model for areal data such as earth observation images, MRI medical images, and county level disease count map. An MRF is a random field that satisfies the Markov property: the conditional probability of the observation at one cell given observations at all remaining cells only depends on observations at its neighbors.  This property is consistent with the first law of geography that ``nearby things are more related than distant things". According to the Brook's lemma~\cite{brook1964distinction}, the joint distribution of cell observations can be uniquely determined based on  conditional probability specified in the Markov property. Furthermore, according to the Hammersley-Clifford theorem~\cite{clifford1990markov}, the corresponding joint distribution of MRF has a unique structure: it can be expressed by a set of potential functions on spatial neighbor cliques (i.e., symmetric functions that are unchanged by any permutations of input variables within a clique). Such a joint distribution is also called Gibb's distribution. Equation~\ref{eq:gibbs} is an example. Its potential function is $W_{i,j}(y(\mathbf{s_i})-y(\mathbf{s_j}))^2$ based on cliques of size two ($\mathbf{s_i},\mathbf{s_j}$). 
The Markov property simplifies the modeling process: as long as the neighborhood structure is specified, the joint distribution of an MRF can be expressed by a potential function on neighbor cliques. Spatial prediction methods based on MRF include ones that explicitly capture spatial dependency such as Simultaneous Autoregressive models (SAR), ones that implicitly capture spatial dependency such as Conditional Autoregressive models, and ones integrating MRF with other models such as Bayesian classifiers and support vector machines.
\begin{equation}\label{eq:gibbs}
    P(y(\mathbf{s_1}),...,y(\mathbf{s_n}))\propto exp\{-\frac{1}{2\sigma^2}\sum_{i,j}W_{i,j}(y(\mathbf{s_i})-y(\mathbf{s_j}))^2\}
\end{equation}

\emph{Simultaneous Autoregressive (SAR) models} (also called \emph{spatial autoregressive models} in spatial econometrics)
explicitly express spatial dependency across response
variables~\cite{anse-88}\cite{banerjee2014hierarchical}. SAR models can be better explained by comparison with linear regression. Classical linear regression model is expressed as Equation~\ref{eq:lr}, where $\mathbf{Y}$ is a $n$ by $1$ column vector 
\begin{equation}\label{eq:lr}
    \mathbf{Y} = \mathbf{X}\boldsymbol{\beta} + \boldsymbol{\epsilon}
\end{equation}
of all response variables, $\mathbf{X}$ is a $n$ by $m$ sample covariate (feature) matrix, $\boldsymbol{\beta}$ is a $m$ by $1$ column vector of coefficients, and $\boldsymbol{\epsilon}$ is a $n$ by $1$ column vector of i.i.d. Gaussian noise (residual errors). In contrast, the SAR model extends traditional linear regression with an additional spatial autoregressive term, as shown in Equation~\ref{eq:sar}, where $\mathbf{W}$ is row-normalized W-matrix, and $\rho$ reflects the strength of spatial
\begin{equation}\label{eq:sar}
    \mathbf{Y} = \rho \mathbf{W} \mathbf{Y} + \mathbf{X} \boldsymbol{\beta} + \boldsymbol{\epsilon}
\end{equation}
dependency effect. The $i$th row of the spatial autoregressive term $\mathbf{W} \mathbf{Y}$ is a weighted average of response variables at all neighboring locations of $\mathbf{s_i}$. Another way to look at SAR is that it multiplies a ``smoother" term $(\mathbf{I}-\rho \mathbf{W})^{-1}$ to the mean and residual error of classical linear regression, i.e., $\mathbf{Y} = (\mathbf{I}-\rho \mathbf{W})^{-1}\mathbf{X} \boldsymbol{\beta} + (\mathbf{I}-\rho \mathbf{W})^{-1}\boldsymbol{\epsilon}$. Parameters in SAR can be estimated based on the maximum likelihood method. SAR model can also be extended for spatial classification via logit transformation. It is worth noting that the spatial autoregressive term can also be added into other variables than the responses, such as covariates as in the spatial Durbin model or residual errors as in the spatial error model~\cite{viton2010notes}.

\emph{Conditional autoregressive (CAR) models} implicitly express spatial dependency via conditional distribution. One common example in the Gaussian case is shown in Equation~\ref{eq:car},
\begin{equation}\label{eq:car}
 y(\mathbf{s_i})|y(\mathbf{s_{j}})_{j\neq i}\sim N(\sum_j \frac{W_{ij}}{W_{i+}}y(\mathbf{s_j}), \frac{\sigma^2}{W_{i+}})   
\end{equation}
where $W_{ij}$ is an element of W-matrix, and $W_{i+}$ is the sum of the $i$th row. In this case, the conditional distribution of a random variable $y(\mathbf{s_i})$ given all other random variables $y(\mathbf{s_{j}})_{j\neq i}$ follows a Gaussian distribution with a mean of neighborhood weighted average. It has been shown in~\cite{banerjee2014hierarchical} that the corresponding joint distribution is the same as Equation~\ref{eq:gibbs}. The main advantage of CAR models is that spatial dependency can be easily captured via potential functions on neighboring cliques. However, the joint distribution can be improper (the integral of the CAR model above is not equal to one). In practice, such CAR models are often used as a prior distribution for Bayesian models~\cite{assunccao2009neighborhood}.

\emph{Integrating MRF with other models}: MRF models can also be integrated with other classification and regression methods to incorporate spatial dependency. One important example is MRF-based Bayes classifiers. Bayes classifiers are classification models that utilize maximum a posteriori probability (MAP) estimate in Bayes theorem, i.e.,
\begin{equation}\label{eq:bc}
\begin{split}
 \widehat{\mathbf{Y}}&=\arg\max_{\mathbf{Y}} \ln P(\mathbf{Y}|\mathbf{X}) \\
 &= \arg\max_{\mathbf{Y}} \ln P(\mathbf{Y}) + \ln P(\mathbf{X}|\mathbf{Y}) \\
 &= \arg\max_{\mathbf{Y}} \sum_i \ln P(y(\mathbf{s_i})) + \sum_i \ln P(\mathbf{x(s_i)}|y(\mathbf{s_i}))   
\end{split}
\end{equation}
where $\mathbf{X}$ and $\mathbf{Y}$ are features and class labels for all samples respectively. The last step of Equation~\ref{eq:bc} is based on the i.i.d. assumption. MRF-based Bayes classifier~\cite{li2009markov} replaces the i.i.d. assumption with spatial dependency via MRF models in order to reduce salt-and-pepper noise~\cite{chawla_modeling_2001,Shekhar-02}.  Specifically, the joint distribution $P(\mathbf{Y})$ is expressed as a Markov random field with potential functions defined on neighbor cliques. For example, 
\begin{equation}\label{eq:mrfbc}
\begin{split}
 \widehat{\mathbf{Y}}&=\arg\max_{\mathbf{Y}} \ln P(\mathbf{Y}|\mathbf{X}) \\
 &= \arg\max_{\mathbf{Y}} \ln P(\mathbf{Y}) + \ln P(\mathbf{X}|\mathbf{Y}) \\
 &= \arg\max_{\mathbf{Y}}~\left\{ \sum_{ij}\lambda W_{ij}(1-\delta(y(\mathbf{s_i}),y(\mathbf{s_j})))\right. \\
 &~~~~~~~~~~~~~~~~~~\left.+ \sum_i \ln P(\mathbf{x(s_i)}|y(\mathbf{s_i}))\right\}
\end{split}
\end{equation}
where $\lambda$ is the weight for spatial dependency, the term $1-\delta(y(\mathbf{s_i}),y(\mathbf{s_j}))$ is a potential function with $\delta$ as Kronecker delta function (i.e., $\delta(y(\mathbf{s_i}),y(\mathbf{s_j}))=1$ if $y(\mathbf{s_i})=y(\mathbf{s_j})$, and $\delta(y(\mathbf{s_i}),y(\mathbf{s_j}))=0$ otherwise). The posterior probability shown in Equation~\ref{eq:mrfbc} can be considered as an energy function, which is the sum of potential functions on neighboring classes $y(\mathbf{s_i})$ and $y(\mathbf{s_j})$ as well as between the feature $\mathbf{x(s_i)}$ and class $y(\mathbf{s_j})$ for each sample. Parameters in MRF-based Bayes classifiers can be estimated via graph cut~\cite{boykov2001fast} and iterations~\cite{besag1986statistical,jackson2002adaptive}. MRF-based models with other potential functions have been applied to earth science problems such as drought detection~\cite{fu_drought_2012}.

Another model similar to MRF is \emph{conditional random field} (CRF)~\cite{lafferty2001conditional}, which directly models spatial dependency within the conditional probability function $P(\mathbf{Y}|\mathbf{X})$ in Equation~\ref{eq:mrfbc}. Its potential function on class labels within a clique is conditioned on feature vector $\mathbf{X}$. Several variants of CRF have been proposed including decoupled conditional random field~\cite{lee_efficient_2006}, discriminative random field~\cite{kumar2003discriminative} and support vector random field~\cite{lee_support_2005}. The difference between MRF and CRF is that the former is generative while the latter is discriminative.

The advantage of MRF-based models is that spatial dependency can be modeled in a very intuitive and simple way (designing potential functions). The limitations include high computational cost in parameter estimation and strong assumptions on the structure of joint probability distribution. In addition, neighborhood relationships are assumed to be given as inputs. Thus, fixed neighborhoods such as square windows are often used for simplicity. For applications where spatial data is anisotropic, determining spatial neighborhood structure is also a challenge.

\emph{(2) Gaussian process based (Kriging)}

Different from MRF based models that are used for areal data, Gaussian process based models are used for spatial prediction (interpolation) on point reference data~\cite{banerjee2014hierarchical}. Given observations of a variable at sample locations in continuous space, the problem aims to interpolate the variable at an unobserved location. Gaussian process assumes that observations at any set of sample locations jointly follow a multivariate Gaussian distribution. The mean term at a location is determined by its local covariates. The residual error term at a location is assumed to be weakly stationary and isotropic, so that the covariance matrix can be expressed as a function of distance (covariogram). 

Specifically, Gaussian process (Kriging) assumes that any set of sample observations $\mathbf{Y}=[y(\mathbf{s_1}),...,y(\mathbf{s_n})]^T$ follows a multivariate Gaussian distribution $N(\boldsymbol{\mu},\boldsymbol{\Sigma})$, where $\boldsymbol{\mu}=\mathbf{X}\boldsymbol{\beta}$ and $\boldsymbol{\Sigma}$ is the covariance matrix $\Sigma_{ij}=Cov(y(\mathbf{s_i}),y(\mathbf{s_j}))=C(\mathbf{s_i}-\mathbf{s_j})$. The main difference of Gaussian process from classical regression is that the residual errors are not mutually independent (the covariance matrix is not diagonal, and the non-diagonal elements can be determined based on covariogram $C(\mathbf{h})$). It can be shown that the optimal predictor (minimizing expected square loss) of $y(\mathbf{s_0})$ given other observations $y(\mathbf{s_1}),...,y(\mathbf{s_n})$ is the conditional expectation as shown in Equation~\ref{eq:gp}, which can be estimated based on the covariance structure from covariogram~\cite{banerjee2014hierarchical}. This method is 
also called \emph{universal} Kriging since it involves covariates $\mathbf{X}$. Special cases without covariates include \emph{simple Kriging} (with known constant mean) and \emph{ordinary Kriging} (with unknown constant mean)~\cite{zimmerman1999experimental}. 
\begin{equation}\label{eq:gp}
    \widehat{y}(\mathbf{s_0})=E[y(\mathbf{s_0})|y(\mathbf{s_1}),...,y(\mathbf{s_n})]
\end{equation} 

\begin{table}
    \centering
    \caption{Comparison between Markov random field and Gaussian process}
    \label{tab:mrfgp}
    \begin{tabular}{p{0.6in}p{1.1in}p{1.2in}}\hline
        Method & Markov random field & Gaussian process \\ \hline
        Spatial data & Areal data & Point reference data\\ \hline
        Spatial dependency & Neighborhood adjacency matrix & Variogram (covariance function on distance)\\ \hline
        Assumption & Conditional independence given neighbors & Isotropy (covariance depends on distance only)\\ \hline
    \end{tabular}
\end{table}

Gaussian process shares some similarities with MRF in that both of them  incorporate spatial dependency (autocorrelation) across sample locations into model structure.  There are several differences, however, as summarized in Table~\ref{tab:mrfgp}: first, Gaussian process is developed for point reference data while MRF is developed for areal data; second, MRF models spatial dependency through W-matrix ($W$) while Gaussian process models spatial dependency through covariance function on spatial distance (covariogram). Similar to MRF which assumes a given neighborhood structure and the Markov property, Gaussian process has assumptions on spatial stationarity and isotropy.

\emph{(3) Hierarchical model with latent variables}

Spatial autocorrelation or dependency can be incorporated into predictive models by adding some spatially autocorrelated latent variables into Bayesian hierarchical models (graphical models). Indeed, the Markov random field based models and Gaussian process models discussed above can be considered as special cases or building blocks for hierarchical models with latent variables.

\begin{figure}[h]
    \centering
    \includegraphics[width=2in]{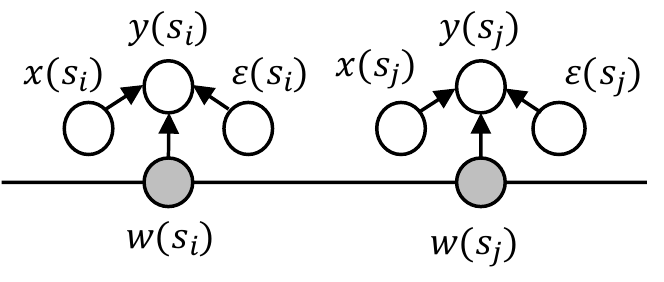}
    \caption{Hierarchical model view for Gaussian process}
    \label{fig:hiegp}
\end{figure}

For instance, in Gaussian process (Kriging), sample observations $\mathbf{Y}$ is assumed to follow a multivariate Gaussian distribution $N(\boldsymbol{\mu}, \boldsymbol{\Sigma})$ where $\boldsymbol{\mu}=\mathbf{X}\boldsymbol{\beta}$, and the covariance matrix $\boldsymbol{\Sigma}$ can be decomposed into two components, one is $\sigma^2\mathbf{I}$ for i.i.d. Gaussian noise, and the other is a non-diagonal matrix $\mathbf{H}$ to capture covariance based on covariogram $H_{ij}=C(\mathbf{s_i}-\mathbf{s_j})$ (introduced in Section~\ref{subsubsec:sa}). This formulation can be rewritten as a hierarchical model in Equation~\ref{eq:hier1}, where $\omega\mathbf{(s_i)}$ is a latent 
\begin{equation}\label{eq:hier1}
    y(\mathbf{s_i}) = \mathbf{x(s_i)}^T\boldsymbol{\beta} + \omega\mathbf{(s_i)} + \epsilon\mathbf{(s_i)}
\end{equation} 
variable and the vector $[\omega\mathbf{(s_1)},...,\omega\mathbf{(s_n)}]^T$ follows a multivariate Gaussian distribution with zero mean and covariance matrix $\mathbf{H}$~\cite{banerjee2014hierarchical}.  The hierarchical model expresses the response variables by adding i.i.d. noise and non-i.i.d. variables for spatial effect. It is illustrated in Figure~\ref{fig:hiegp}, in which $\epsilon\mathbf{(s_i)}$ and $\epsilon\mathbf{(s_j)}$ (as well as $\mathbf{x(s_i)}$ and $\mathbf{x(s_j)}$) are independent, $y(\mathbf{s_i})$ and $y(\mathbf{s_j})$ are conditionally independent given latent variables $\omega\mathbf{(s_i)}$ and $\omega\mathbf{(s_j)}$. Latent variables $\omega\mathbf{(s_i)}$ at different locations $\mathbf{s_i}$ reflect the spatial (autocorrelation) effect since they follow a joint Gaussian distribution with a non-diagonal covariance matrix (i.e., correlation exists across neighboring $\omega\mathbf{(s_i)}$ and $\omega\mathbf{(s_j)}$). 

\begin{figure}[h]
    \centering
    \includegraphics[width=1.92in]{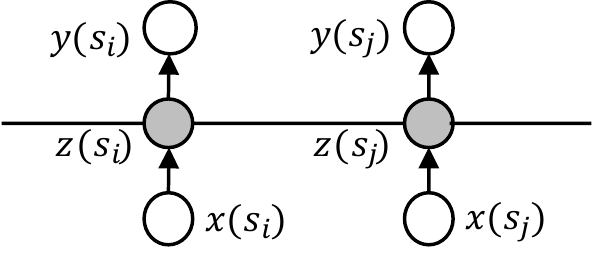}
    \caption{Hierarchical model view for spatial sequence with corrupted class labels $y$}
    \label{fig:hier2}
\end{figure}

Another instance of example is disease risk mapping on areal data~\cite{banerjee2014hierarchical}. The count of disease events at one county $\mathbf{s_i}$ can be assumed to follow a Poisson distribution $Po(n(\mathbf{s_i}) r(\mathbf{s_i}))$ where $n(\mathbf{s_i})$ is the known number of high risk people, $r(\mathbf{s_i})$ is the variable for disease risk ($r(\mathbf{s_i})\in[0,1]$), and $n(\mathbf{s_i}) r(\mathbf{s_i})$ is the expectation of the Poisson distribution in county $\mathbf{s_i}$. We assume the distribution in Equation~\ref{eq:hier2a}, where $\mathbf{x(s_i)}$ is the covariate vector, $\boldsymbol{\beta}$ is the  
\begin{equation}\label{eq:hier2a}
    r(\mathbf{s_i})=exp\{\mathbf{x(s_i)}^T\boldsymbol{\beta} + \epsilon(\mathbf{s_i})\}
\end{equation}
coefficient vector, and $\epsilon(\mathbf{s_i})$ is i.i.d. Gaussian noise, then the problems becomes a non-spatial problem, i.e., each county is independent from each other. However, in reality, though disease risk at different counties may be different, nearby counties have a high tendency to have similar risks. To reflect this phenomena, we can further model the disease risks at different counties as in Equation~\ref{fig:hier2}, where $\omega\mathbf{(s_i)}$ is an  
\begin{equation}\label{eq:hier2}
 r(\mathbf{s_i})=exp\{\mathbf{x(s_i)}^T\boldsymbol{\beta} + \omega\mathbf{(s_i)} + \epsilon(\mathbf{s_i})\}   
\end{equation}
additive term for the spatial autocorrelation effect. For instance, $\omega\mathbf{(s_i)}$ can be a conditional autoregressive (CAR) model. This hierarchical model has a similar structure as the one in Figure~\ref{fig:hiegp} based on the similar conditional independence assumption.

Latent variable can also be used to incorporate spatial autocorrelation for corrupted observation data. Observations of a target variable may be corrupted or missing due to noise or errors in data collection process, even though the true values of the variable should be uncorrupted with a high spatial autocorrelation. Corrupted observations will impact the performance of predictive models since training samples are inaccurate. To address this issue, a latent variable approach has been proposed~\cite{DBLP:conf/kdd/KimYTM15}, in which a corrupted observation $y(\mathbf{s_i})$ depends on a corresponding uncorrupted (spatially autocorrelated) latent variable $z(\mathbf{s_i})$, i.e., $y(\mathbf{s_i})=f(z(\mathbf{s_i}))$, and the latent variable depends on covariates $z(\mathbf{s_i})=g(\mathbf{x(s_i)})$. The entire model is hierarchical as illustrated in Figure~\ref{fig:hier2}. Model learning involves estimating parameters for functions $f$ and $h$.

The advantage of using hierarchical models with latent variables to incorporate spatial dependency is that the modeling process is simple and intuitive, providing flexibility in model design. For example, such models have been used in real estate appraisal~\cite{DBLP:conf/kdd/FuXGYZZ14} and event forecasting from social media data~\cite{DBLP:conf/sdm/ZhaoCLR15}.
The main issue is the computational cost. Learning parameters involves iterative methods such as EM algorithms or Markov Chain Monte Carlo simulation. The computational cost can be high for large data with many nodes (variables).

\subsubsection{Spatial regularization in objective function}\label{sect:sasr}
In addition to explicitly modify model structure to capture spatial dependency, we can also extend the objective (or loss) function with an additional spatial regularization term. In this way, the learning algorithm favors parameter values that not only make accurate prediction at individual locations but also show high spatial autocorrelation in predicted map. The model here can refer to a single model such as linear regression or a composite of multiple models with one at each location.

\begin{table*}
    \centering
    \caption{Comparison of different methods addressing spatial autocorrelation}
    \label{tab:compauto}
    \begin{tabular}{p{1.2in}p{2.5in}p{2.5in}}\hline
     Method &Advantages& Disadvantages\\ \hline
     Spatial contextual feature generation & Easy to use, do not require modifying models and learning algorithms, no restriction on model types & Subjective, need to regenerate all candidate features for each problem \\ \hline
     Spatial dependency in model structure & Intuitive, clear theoretical properties  & Restricted to fixed type of models and distributions, model learning is computational expensive\\ \hline
     Spatial regularization in object function & Intuitive, clear theoretical properties, less restriction on types of models & Requiring models with differentiable objective functions, no guarantee with optimal solutions, computationally expensive\\ \hline
    \end{tabular}
\end{table*}


\emph{Spatial regularization in multi-model prediction:} Multi-model prediction utilizes a composite of local models with one model at each location (or sub-region) in the study area. Each local model has its own parameters that can be learned by minimizing prediction errors on learning samples at this location. However, independently learning these local models at individual locations risks overfitting due to the large number of parameters in multiple models. To solve this issue, spatial autocorrelation constraint on model parameters can be added to the objective function. The main idea is to create an overall loss function by summing up individual loss of local models  with a regularization term that penalizes inconsistent model parameters at neighboring locations. The underlying assumption is that nearby models should have similar parameters due to spatial autocorrelation.

One common spatial regularizer based on the spatial autocorrelation effect is graph Laplacian regularizer~\cite{weinberger2007graph}. The idea is to consider each local model as a node and spatial neighborhood relationships between locations as edges, and to penalize neighboring nodes with very different parameters. Approaches have been proposed with different types of base models, including linear regression~\cite{subbian_climate_2013}, logistic regression~\cite{DBLP:conf/sdm/DataKKBK14}, and support vector machine~\cite{stoeckel_svm_2005,DBLP:conf/sdm/DataKKBK14}. Consider linear regression base model as an example, the traditional loss function is the sum of square errors, i.e., 
\begin{equation}
L(\boldsymbol{\Theta})=\sum_i \|\mathbf{X(s_i)}\boldsymbol{\theta}\mathbf{(s_i)}-\mathbf{Y(s_i)}\|_2^2
\end{equation}
where $\mathbf{s_i}$ is the location (or region) for the $i$th model (task), $\mathbf{X(s_i)}$ and $\mathbf{Y(s_i)}$ are the feature matrix and response vector for all samples at $s_i$, and $\boldsymbol{\Theta}=[\boldsymbol{\theta}\mathbf{(s_1)},...,\boldsymbol{\theta}\mathbf{(s_n)}]$ is a $m$ by $n$ matrix of all model parameters. Minimizing $L(\boldsymbol{\Theta})$ is equivalent to minimize loss functions on each location $s_i$ independently since there parameters $\boldsymbol{\theta}\mathbf{(s_i)}$ and $\boldsymbol{\theta}\mathbf{(s_j)}$ are independent for $i\neq j$. This leads to a large number of parameters and potential risks for model overfitting. To address this issue, graph Laplacian regularizer is added, penalizing differences of model parameters at neighboring locations. Specifically, the regularization term is 
\begin{equation}
\Omega(\boldsymbol{\Theta})=\frac{1}{2}\sum_{ij}W_{ij}\|\boldsymbol{\theta}\mathbf{(s_i)}-\boldsymbol{\theta}\mathbf{(s_j)}\|_2^2=Trace(\boldsymbol{\Theta} \mathbf{L} \boldsymbol{\Theta}^T)   
\end{equation}
where  $W_{ij}$ is the element of W-matrix with $W_{ij}=1$ when $\mathbf{s_i}$ and $\mathbf{s_j}$ are neighbors and $W_{ij}=0$ otherwise, $\mathbf{L}$ is a graph Laplacian matrix. 
The combined loss function will be Equation~\ref{eq:sploss}, where $\lambda$ controls the degree of spatial smoothness among parameters. 
\begin{equation}\label{eq:sploss}
    L(\boldsymbol{\Theta})=\sum_i \|\mathbf{X(s_i)}\boldsymbol{\theta}\mathbf{(s_i)}-\mathbf{Y(s_i)}\|_2^2+\lambda Trace(\boldsymbol{\Theta} \mathbf{L} \boldsymbol{\Theta}^T)
\end{equation} 

The advantage of this approach is that the model is very intuitive, easily interpretable, and generally applicable to different base models as long as their loss function is differentiable. In practice, parameters can be estimated iteratively via Newton Raphson methods~\cite{avriel2003nonlinear}. It is worth noting that the graph Laplacian regularizer here has subtle differences from the one often used in semi-supervised learning. In semi-supervised learning, there is only one model instead of multiple models, and the regularizer penalizes parameter values whose model predictions at neighboring locations are inconsistent.

\emph{Spatial decision trees:} Decision tree classifiers have been widely used for spatial classification problem in earth science~\cite{hansen2000global,pal2003assessment} due to simplicity, interpretability, computational efficiency, and being non-parametric. However, decision tree implicitly assumes that samples are independent and identically distributed. This assumption is often violated in spatial data due to spatial autocorrelation, resulting in artifacts in prediction results. To address this limitation, \emph{spatial decision trees} have been proposed~\cite{jiang_learning_2012,li_spatial_2006,stojanova2011global,stojanova2012dealing,jiang2015focal} that incorporate the spatial autocorrelation effect into decision tree learning algorithms. This is usually done by modifying the entropy or information gain heuristic. The main idea is that selection of a tree node test should be based on not only class purification but also the spatial arrangement of samples being split on the map.

Specifically, decision trees often use entropy and information gain measures to select tree node tests. Entropy measures the impurity of class distribution. Its value is high when the probabilities of different classes are close with each other (high class impurity). Information gain is defined as the decrease of entropy after training samples are split by a tree node test. Decision tree learning algorithms select a tree node test with the maximum information gain each time. However, this heuristic ignores the spatial pattern on how training samples are split on a map. According to spatial autocorrelation, we can assume that nearby training samples with the same class should be split into the same subset, such that they are likely to be predicted into the same class. Several approaches have been proposed that extend the traditional entropy or information gain definition with spatial regularization, including spatial entropy based on spatial distance~\cite{li_spatial_2006}, spatial information gain based on spatial autocorrelation statistics such as Moran's I and Geary's C~\cite{stojanova2011global,stojanova2012dealing} and Gamma index~\cite{jiang_learning_2012}. 
Different spatial entropy or information gain definitions make different assumptions on the ground truth class map. Distance based spatial entropy assumes that samples from each class form a globally compact cluster with high inter-class sample distance and low inner-class sample distance. Spatial autocorrelation statistics based spatial information gain assumes that there are good tree node tests that can separate neighboring samples from different classes into different subsets while keeping neighboring samples from the same class within the same subset.

Spatial decision tree inherits the merits of decision tree model family such as interpretability and non-parametric nature. However, since entropy and information gain is only a greedy heuristic, there is no guarantee on global optimality. For example, if all input feature maps tend to have poor spatial autocorrelation, spatial entropy or information gain will still select one among them. In this scenario, extending the candidate tree node tests to incorporate neighborhood autocorrelation statistics will help~\cite{jiang2015focal}.

{\it Spatial accuracy objective function:} In traditional classification problems, the objective function is often measured on each sample, e.g., if a sample is misclassified or not, regardless how far the predicted class is away from the nearest true class. Consider the example of classifying raster cells into ``with bird nest" and ``without bird nest". If a cell that is mistakenly predicted as  ``with bird nest" is very close to an actual bird nest cell, the prediction accuracy on this sample should be considered higher than zero. Thus, spatial accuracy~\cite{chawla_modeling_2001} has been proposed to measure not only  how accurate each cell is predicted by itself but also how far it is from the nearest location of its true class. 

Table~\ref{tab:compauto} compares the three different strategies to incorporate spatial autocorrelation (or dependency) into spatial prediction, including spatial contextual feature generation, spatial dependency within model structure, and spatial regularization in objective function. Spatial feature generation is simple and intuitive, without the need of modifying model structures and learning algorithms. But selecting candidate features can be subjective, requiring strong knowledge related to the application domain. The process needs to be repeated for a different problem instance. The latter two approaches are intuitive with clear theoretical properties, making it easy to design models and learning algorithms. But they rely on strong assumptions on model types, and are often computationally expensive. 

\subsection{Spatial Heterogeneity}\label{sect:sh}
Spatial heterogeneity is another major challenge in spatial prediction problems. Spatial data samples often do not follow an identical distribution in the entire study area. A global model that is learned from samples in the entire study area may produce poor predictions for local regions. For example, the relationships between house price and house age may differ dramatically between suburb and urban areas. There are several approaches in the literature to address the challenge of spatial heterogeneity, including location dependent model parameters, decomposition based spatial ensemble, and multi-task learning. The main idea behind these methods is to use a composite of local or regional models that are location sensitive to replace a single global model.

\subsubsection{Spatial coordinate features}

One simple strategy to make a model location sensitive is to incorporate spatial coordinates into the feature (covariate) vector. For example, incorporating spatial coordinate features into regression can fit a trend surface that is location dependent. Similarly, spatial coordinate features in decision trees can split samples not only in the feature space but also in the geographic space. However, due to the fact that spatial locations are multi-dimensional, considering spatial coordinates as separate features cannot effectively model heterogeneous spatial data where homogeneous sub-regions have arbitrary footprint shapes. For example, a decision tree model learned from data with spatial coordinate features often partitions the geographical space in parallel with the vertical or horizontal axis, and thus cannot effectively partition the geographic space into irregular footprints.

\begin{table*}
    \centering
    \caption{Comparison of different methods addressing spatial heterogeneity}
    \label{tab:comphetero}
    \begin{tabular}{p{1.2in}p{2.5in}p{2.5in}}\hline
     Method &Advantages& Disadvantages\\ \hline
     Geographical weighted model & Simple and intuitive, clear theoretical properties, no restriction on types of models & Underlying assumption of isotropy can be invalid\\ \hline
     Decomposition based spatial ensemble &No restriction on types of models and shapes of homogeneous zones &Decomposing space into homogeneous zones for local models can be non-trivial \\ \hline
     Multi-task learning & No restriction on shapes of homogeneous zones, capturing spatial dependency between local models & Restriction to local models with differentiable objective functions, decomposing space into homogeneous zones for local models can be non-trivial\\ \hline
     
    \end{tabular}
\end{table*}

\subsubsection{Geographically Weighted Models} 
A geographically weighted model addresses the challenge of spatial heterogeneity, particularly spatial non-stationarity, by learning a distinct model at each location so that the model parameters are location dependent. When learning a local model, training samples are geographically weighted so that nearby samples have higher weights. 

\begin{figure}
    \centering
    \includegraphics[width=3.5in]{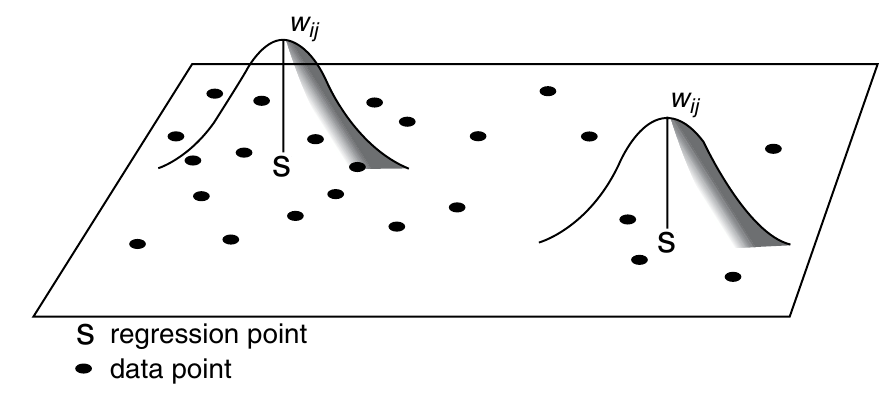}
    \caption{Illustration of geographically weighted model based on kernel function (source of image:\cite{fotheringham2002geographically})}
    \label{fig:gwr}
\end{figure}

One common example is geographically weighted regression (GWR)~\cite{fotheringham2002geographically}. GWR can be better explained through comparison with classical linear regression. In linear regression, $y(\mathbf{s_i})=\mathbf{x(s_i)}^T\boldsymbol{\beta} + \epsilon(\mathbf{s_i})$, model coefficients $\boldsymbol{\beta}$ are assumed to be identical for the entire study area. In contrast, the model coefficients of GWR is location dependent $y(\mathbf{s_i})=\mathbf{x(s_i)}^T\boldsymbol{\beta}(\mathbf{s_i}) + \epsilon(\mathbf{s_i})$. The coefficient
$\boldsymbol{\beta}(\mathbf{s_0})$ at a new location $\mathbf{s_0}$ can be learned by weighted least square errors, where the weight for each training sample is determined by its distance to $\mathbf{s_0}$. Specifically, this is shown in Equation~\ref{eq:gwr},
\begin{equation}\label{eq:gwr}
    \beta(\mathbf{s_0})=\mathrm{argmin}_{\beta(\mathbf{s_0})} \sum_i w(\mathbf{s_i},\mathbf{s_0})(y(\mathbf{s_i})-\mathbf{x(s_i)}^T\boldsymbol{\beta(s_0)})^2
\end{equation}
where $w(\mathbf{s_i},\mathbf{s_0})$ is determined by a spatial kernel weighting function, e.g., $w(\mathbf{s_i},\mathbf{s_0})=exp\{-\frac{1}{2}\|\mathbf{s_i}-\mathbf{s_0}\|_2^2\}$. A sample that is closer to the current model location has a higher weight following the spatial autocorrelation effect, as illustrated by Figure~\ref{fig:gwr}. It is worth noting that both GWR and SAR (Section~\ref{sec:sainm}) are spatial regression models that incorporates spatial autocorrelation or dependency. The main difference is that SAR still assumes that model coefficients are same for all locations, and thus does not address the challenge of spatial heterogeneity, while GWR addresses spatial heterogeneity by learning a set of model parameters at each location. 

The main idea of geographically weighted models using spatial kernel weighting has been extended to general linear models~\cite{nakaya2005geographically} and principle component analysis~\cite{harris2011geographically}. It can also be generalized to many other methods, such as decision trees, support vector machines. In these cases, we only need to precompute the relative weights for all other samples to a location of interest so that a local model can be learned for that location. 

The advantages of geographically weighted models include simplicity and effectiveness in incorporating local effects. Two main limitations exist though. First, the computational cost is high since a model needs to be learned for each location of interest in the continuous space. Second, geographical weights of samples are determined by spatial kernel functions assuming that the geographical influence of samples to a location only depends on their distances (spatial isotropy). This assumption is often violated by real world geographic data due to spatial anisotropy when homogeneous sub-regions have arbitrary shape. 

\subsubsection{Decomposition based ensemble}
The assumption is that spatial data consists of a number of homogeneous sub-population within which the relationship between sample features and class labels is consistent. Based on this assumption, decomposition based ensemble learning uses a divide-and-conquer strategy to first partition data into different homogeneous sub-groups, and then learns a local model in each sub-group. This approach generally belongs to ensemble learning~\cite{ren2016ensemble,zhou2012ensemble,dietterich2000ensemble}, which aims to boost predictive accuracy by learning multiple based models. 

Several decomposition based ensemble methods partition multi-modular input data in feature vector space, including
mixture of experts~\cite{jacobs1991adaptive,jordan1994hierarchical}  and multimodal ensemble~\cite{KarpatneK15,karpatne2015ensemble}.
Partitioning is often done via feature clustering, or a gating network. However, partitioning input data in feature vector space may not effectively separate samples with class ambiguity, i.e., samples with similar explanatory features belong to different classes in different spatial regions. Several approaches have been proposed to partition spatial samples in the geographical space. One approach is to use auxiliary information such as road networks and census blocks together with spectral information to segment  satellite imagery into different sub-regions~\cite{vatsavai_hybrid_2011}.  Similarly, a two-step spatial ensemble method has been proposed that first segments sample locations into homogeneous patches, then group patches into different zones via a bisecting algorithms~\cite{jiang_ensemble_2017}.  Another approach is based on competition strategy~\cite{radosavljevic_spatio-temporal_2008}, which involves an initial partitioning, and sample shifting across boundaries. 

Decomposition based ensemble has several advantages. First, the spatial footprints of local models can have arbitrary shapes. This overcomes the limitations of geographically weighted models, which assume spatial isotropy (circular shapes).  Second, once spatial decomposition is done in preprocessing, traditional classification models can be used for each sub-region without the need of developing new classification algorithms. Several limitations exist as well. First, finding a good decomposition in geographic space is computationally challenging. Second, most decomposition-based spatial ensemble approaches assume that future test samples lie within the same spatial framework as training samples. Thus, methods cannot be applied to test samples beyond the training area.

\subsubsection{Multi-task learning}
Multi-task learning is a common machine learning technique for heterogeneous data~\cite{Caruana1997}. The main idea is to group learning samples into different tasks, and to simultaneously learn models in different tasks according to task relatedness (e.g., models from related tasks share the same set of features). When used to address spatial heterogeneity, multi-task learning approach is similar to decomposition based ensemble approach above in that it also learns local models (tasks) in different regions or locations, but it is different in that parameters of local models are jointly learned together according to the task relatedness (nearby models tend to have similar parameters). 

The main question is how to identify different tasks and task relationships. One approach is to use each location (spatial point or raster cell) as a task and to use spatial neighborhood relationships as task relatedness, which can be determined by spatial distance or cell adjacency~\cite{subbian_climate_2013,DBLP:conf/kdd/ZhaoSYCLR15}.
Another approach is to conduct spatial clustering on sample locations. Each cluster is a task, and clusters that are closer to each other are more related~\cite{DBLP:conf/sdm/DataKKBK14}. Task relatedness can also be determined by inferring conditional independence of class probability distribution~\cite{gonccalves2015multi}. When learning models from related tasks, a graph Laplacian regularizer can be used to enforce that nearby models have similar parameters~\cite{subbian_climate_2013,DBLP:conf/sdm/DataKKBK14} (details are in spatial regularization for multiple model prediction in Section~\ref{sect:sasr}). Graph Laplacian regularizer will not only incorporate spatial autocorrelation effect, but also reduce overfitting when the number of  training samples is limited. The base models can be linear model, or generalized linear model such as logistic regression, as well as other models as long as the objective function is differentiable. 

Similar to decomposition based ensemble, multi-task learning approach has the advantage of flexibility in spatial footprint shapes of sub-regions. It has additional advantages of incorporating spatial autocorrelation and avoiding overfitting with limited training samples. However, it also has more constraints on the choice of base models (those with differentiable objective functions). In addition, determining sub-regions for local tasks as well as relationships between tasks can be non-trivial. A detailed comparison of geographically weighted models, decomposition based spatial ensemble, and multi-task learning approaches are summarized in Table~\ref{tab:comphetero}.



\subsection{Limited Ground Truth}\label{subsec:limited}

In real world spatial prediction problems, input data often contains abundant explanatory features but very limited ground truth. For example, in earth image classification for land cover mapping, a large number of learning samples (image pixels) are collected with explanatory features (spectral band values) but only a small set of these samples have ground truth (land cover types). Collecting ground truth is both expensive and time consuming, requiring to send a field crew on the ground or recruit visual interpreters. There are two general strategies to address the challenge, including semi-supervised learning and active learning.

\subsubsection{Semi-supervised learning}
Semi-supervised learning methods utilize labeled samples together with unlabeled samples in model learning to improve prediction performance. There are many different semi-supervised learning methods, including generative models such as mixture of Gaussian with EM algorithm, graph-based methods, transductive support vector machine, co-training and self-training (more details can be found in a survey~\cite{zhu2005semi}). Due to space limit, we briefly introduce three popular methods that have been used in spatial classification and prediction problems.

\emph{Generative mixture of models and EM algorithm:} This method assumes that feature values of samples from different classes follow a mixture model such as mixture of Gaussian. Unlabeled samples are used to improve the estimate of conditional distribution of feature values under each class.
For example, in Gaussian mixture model, the joint distribution of features and classes is $P(\mathbf{x(s_i)},y(\mathbf{s_i})=c_j)=\alpha_{c_j} N(\mathbf{x(s_i)};\boldsymbol{\mu}_{c_j},\boldsymbol{\Sigma}_{c_j})$, where $\alpha_{c_j}$ is the prior probability $P(y(\mathbf{s_i})=c_j)$, $\boldsymbol{\mu}_{c_j}$ and $\boldsymbol{\Sigma}_{c_j}$ are the mean and covariance of the normal distribution for conditional probability $P(\mathbf{x(s_i)}|y(\mathbf{s_i})=c_j)$. 
Unlabeled samples can be incorporated in the maximum likelihood estimation of model parameters. Specifically, the log likelihood function with both labeled and unlabeled samples can be written as
\begin{equation}\label{eq:semi}
\begin{split}
    LL(\mathbf{X},\mathbf{Y};\boldsymbol{\Phi})=\sum_{1\leq i\leq n_L} \log\left(\alpha_{y(\mathbf{s_i})} N(\mathbf{x(s_i)};\boldsymbol{\mu}_{y(\mathbf{s_i})},\boldsymbol{\Sigma}_{y(\mathbf{s_i})})\right)+\\
    \sum_{n_L+1\leq i\leq n_L+n_U} \log\left(\sum_{y(\mathbf{s_i})\in \mathcal{C}}\alpha_{y(\mathbf{s_i})} N(\mathbf{x(s_i)};\boldsymbol{\mu}_{y(\mathbf{s_i})},\boldsymbol{\Sigma}_{y(\mathbf{s_i})})\right)
\end{split}
\end{equation}
where $\boldsymbol{\Phi}$ represents all parameters, $n_L$ and $n_U$ are the numbers of labeled and unlabeled samples respectively. The log likelihood of unlabeled samples are incorporated in the second term of Equation~\ref{eq:semi}, where  their unknown class labels (latent variables) are marginalized out (summation over $y(\mathbf{s_i})\in \mathcal{C}$). Expectation and Maximization (EM) algorithm can be used~\cite{dempster1977maximum} to iteratively update model parameters and the hidden classes of unlabeled samples (latent variables). 

Gaussian mixture model has been used in semi-supervised classification of earth imagery with the maximum likelihood classifiers~\cite{DBLP:conf/icdm/VatsavaiSB08,DBLP:conf/igarss/VatsavaiBSB08}. Its advantages include clear assumptions and theoretical properties. The main limitation is that the assumptions of i.i.d. distribution and Gaussian class conditional distribution can be violated by real world spatial data. Ideas have been explored that incorporate the Markov property into semi-supervised  max likelihood classifiers~\cite{DBLP:journals/paapp/VatsavaiSB07}.  

\emph{Graph-based methods:} The graph-based approach first constructs a graph, in which nodes are spatial data samples (both labeled and unlabeled) and the edges are determined by the distance or similarity between samples (e.g., feature similarity, geographical proximity, or both via composite kernels~\cite{camps2007semi}). The main assumption is that samples that are close with each other have a high chance to share the same class labels. The loss function thus consists of two components, one for the prediction accuracy on labeled samples, and the other for the smoothness of predictions on neighboring samples. Specifically,
\begin{equation}
 L(\mathbf{Y},\widehat{\mathbf{Y}})=\sum_i (y(\mathbf{s_i})-\widehat{y}(\mathbf{s_i}))^2 + \lambda\sum_{i,j} W_{ij}(\widehat{y}(\mathbf{s_i})-\widehat{y}(\mathbf{s_j}))^2
\end{equation}
where $y(\mathbf{s_i})$ and $\widehat{y}(\mathbf{s_i})$ are the true response and predicted response for the sample at location $\mathbf{s_i}$, $W_{ij}$ is the element of W-matrix corresponding to $\mathbf{s_i}$ and $\mathbf{s_j}$ whose value can be determiend by feature similarity or spatial proximity. Various computational algorithms exist to estimate class labels that minimize the loss function, including graph min-cut algorithm, iterations with neighborhood updates, or a closed form solution when the loss function is differentiable~\cite{zhu2005semi}.  
Graph regularizer has also been used with support vector machines for hyperspectral earth image classification~\cite{gomez2008semisupervised}.

The graph-based method looks similar to the graph Laplacian regularization method in Section~\ref{sect:sasr}. The main difference is that in semi-supervised learning, there is only one model and the graph regularizer smoothes model predictions on nearby samples, while in multiple model prediction  in Section~\ref{sect:sasr}, there are multiple models at different locations and the graph regularizer smoothes the parameters of nearby models.
The advantage of graph-based methods for semi-supervised spatial prediction is its simplicity and intuitiveness. However, its performance may be degraded if the assumption on the smoothness of neighboring classes is violated.

\emph{Self-training and co-training:} Self-training methods iteratively expand the set of training samples by adding predicted classes on unlabeled samples. Predictions with the highest confidence will be selected and added into the training set. The model is iteratively updated based on the expanded training set. In spatial classification, spatial neighborhood expansion is often used to select unlabeled samples, i.e., we can select unlabeled samples that are spatial neighbors of a labeled sample~\cite{dopido2013semisupervised,tan2015novel}. This follows the first law of geography, i.e., nearby samples tend to resemble each other in their class labels. Co-training is another semi-supervised learning method. Similar to self-training, it augments the training set by selecting samples with the highest confidence from model prediction. The difference is that co-training uses multiple conditionally independent feature sets (or views) and learns one classifier for each feature set (or view). Each classifier selects an unlabeled sample with highest prediction confidence and adds it to the shared expanded training set. Co-training has been used for spatial classification problems, particularly earth image classification. In this case, different feature sets (views) can be spectral versus spatial features~\cite{hong2015spatial}, derived features from different image sub-blocks~\cite{zhang2014modified}, or features from different data sources with various resolutions such as Landsat~\cite{landsat}, MODIS~\cite{modis}, and high resolution aerial photos. 

The advantages of self-training and co-training include that the method is general for many different model families and that the process is automated without need for extra human intervention. The limitation is that the methods rely on accurate model predictions. If a model prediction with the highest confidence is still erroneous, adding the predicted samples into training sets can further impact model performance. In addition, the computational cost is high since a model needs to be re-trained for each iteration.

\subsubsection{Active learning}
Active learning~\cite{settles2010active,tuia2011survey} addresses the challenge of limited ground truth by manually labeling a carefully selected subset of unlabeled samples. The main question is how to select the subset of unlabeled samples that can enhance prediction performance the most and with the minimum labeling costs. There are several strategies to do this, including selecting samples with the highest uncertainty, samples on which a committee of models disagree the most, samples that have the highest expected model change, or samples with the best expected error reduction. The process is often iterative: models are re-trained after new training samples are added. More details can be found in a survey~\cite{settles2010active}. 

For spatial prediction, sample locations need to be considered when selecting unlabeled samples because collecting class labels often involves sending a field crew traveling between locations on the ground. If selected locations of unlabeled samples are spatially disperse, the travel costs will be high. Thus, unlabeled samples that are spatially clustered are preferred over those that are spatially disperse. To this end, a region-based active learning method~\cite{stumpf2014active} has been proposed for remote sensing image classification. In each iteration, the method selects a window of pixels that have the most overall disagreement by a committee of models. Another spatial cost-sensitive active learning~\cite{liu_spatially_2009}, in order to find a travel route that covers the top-K most uncertain samples with the minimum travel cost.

The advantage of active learning approach is that it can improve labeling efficiency since samples with the most uncertainty are first labeled. Its main limitation is that human labor is still needed so the amount of ground truth data being collected is often limited. In addition, human experts who manually collect ground truth labels may need to wait for model re-training in each iteration.

\subsection{Multiple Scales and Resolutions} 
Another challenge in spatial prediction is that spatial data may exist in multiple spatial scales and resolutions. For example, resolutions of earth observation imagery range from sub-meter to hundreds of meters. In addition to raster imagery, spatial data can also contain point reference data such as soil samples. Many existing predictive models assume that data samples are from the same scale and resolution, and thus are not directly applicable to multi-scale and multi-resolution data. Moreover, spatial patterns or relationships between explanatory features and the target response variable can be scale dependent (also called the Modifiable Area Unit Problem (MAUP)~\cite{wong2009modifiable}). 

One existing strategy to address this challenge is to build spatial hierarchical models. A hierarchical model has multiple layers ordered by spatial scales, and each layer consists of spatial units at the same scale or resolution. Each spatial unit in a layer has its own predictive model, and relationships between parameters in different models can be established based on the scale hierarchy. 
As a specific example, a hierarchical multi-source feature learning framework~\cite{DBLP:conf/kdd/ZhaoYCLR16} has been proposed to forecast spatiotemporal events based on features from multiple scales including cities, states, and countries.  Another example is the problem of modeling count of caries in human teeth. Each subject (person) has a number of tooth, and a tooth has different surfaces. Spatial neighborhood relationships exist across nearby tooth surface in the same subject. A Bayesian hierarchical model has been used to capture such spatial hierarchy~\cite{bandyopadhyay2009bayesian}. 
Spatial hierarchical models have also been used in crime event forecasting~\cite{yu2016hierarchical}. In this work, distributed spatio-temporal patterns from multi-resolution data are ensembled in predictive modeling.


Hierarchical spatial models have several advantages, including intuitive model design, utilizing data from different sources in different scales or resolutions, and capturing spatial dependency within and across spatial scales. The main limitation lies in model complexity, e.g., high computational cost, and risk of overfitting.

\subsection{Spatiotemporal Extensions to Prediction Methods}\label{sec:stprediction}
In many spatial prediction problems, both space and time are critically important. For example, the time of day (e.g., rush hour versus non-rush hour) plays an important role in air quality prediction due to its impacts on the amount of traffic emissions. Adding the time dimension into spatial data creates new  data representation and statistical concepts. More details can be found in a recent survey on spatiotemporal data mining~\cite{shekhar2015spatiotemporal}.

Compared with spatial prediction, spatiotemporal prediction poses two unique challenges: spatiotemporal autocorrelation and temporal non-stationarity. The effect of \emph{spatiotemporal autocorrelation} means that samples tend to resemble each other not only in nearby locations but also at close times. The effect of \emph{temporal non-stationarity} means that statistical properties are dynamic over time. Addressing these challenges require extensions to existing spatial prediction methods. Due to space limit, we only introduce a few well-known examples.

\subsubsection{Spatiotemporal Autocorrelation}
\emph{Spatiotemporal contextual features}: Many methods discussed in Section~\ref{subsubsec:scf} that generate spatial contextual features can be readily used or extended for spatiotemporal data. For example, additional contextual features can be generated in temporal sequences of raster data (e.g., earth imagery) based on sample attributes in spatiotemporal neighborhoods~\cite{mennis2005cubic}. Contextual features can also be generated by fusing data from multiple sources into a common spatiotemporal framework (e.g., spatial grid cells and time intervals)~\cite{zheng2015methodologies}. After features are generated, we can apply traditional non-spatiotemporal prediction methods.

\emph{Spatial panel data model}: Spatial panels refer to spatiotemporal data containing time series observations at a number of spatial locations (e.g., zip codes, cities, states)~\cite{fischer2009handbook}. Spatial panel data models, therefore, refer to prediction models whose response variables are spatial panels. One simple form is a pooled linear regression model with spatial specific effects but without spatial interaction effects, 
\begin{equation}\label{eq:spdm1}
    y(\mathbf{s_i},t)=\mathbf{x}(\mathbf{s_i},t)^T\boldsymbol{\beta}+\mu(\mathbf{s_i})+\epsilon(\mathbf{s_i},t)
\end{equation} where $i\in\mathbb{N},1\leq i\leq n$, $t\in\mathbb{N},1\leq t\leq T$, $n$ and $T$ are the numbers of locations and time steps respectively, $\mu(\mathbf{s_i})$ is the spatial specific term (depending on location regardless of time), and $\epsilon(\mathbf{s_i},t)$ is i.i.d. Gaussian noise. The model can also be represented in matrix form,
\begin{equation}\label{eq:spdm2}
    \mathbf{Y}(t) = \mathbf{X}(t)\boldsymbol{\beta}+\boldsymbol{\mu}+\boldsymbol{\epsilon}(t)
\end{equation}
where $\mathbf{Y}(t)=[y(\mathbf{s_1},t),...,y(\mathbf{s_n},t)]^T$ is a $n$ by $1$ response vector at time $t$, $\mathbf{X}(t)=[\mathbf{x}(\mathbf{s_1},t),...,\mathbf{x}(\mathbf{s_n},t)]^T$ is a $n$ by $m$ covariate (feature) matrix at time $t$,  $\boldsymbol{\mu}$ is a $n$ by $1$ vector for spatial specific effect, and $\boldsymbol{\epsilon}(t)$ is a $n$ by $1$ vector with i.i.d. Gaussian residual errors. Based on the simple form, spatial autoregressive (interaction) term (Section~\ref{sec:sainm}) can be added to the response vector $\mathbf{Y}(t)$ (Equation~\ref{eq:spdm2}), or to the covariate matrix $\mathbf{X}(t)$ (Equation~\ref{eq:spdm3}), or to residual errors (Equation~\ref{eq:spdm4} with $\boldsymbol{\phi}(t) = \rho\mathbf{W}\boldsymbol{\phi}(t)+\boldsymbol{\epsilon}(t)$).
\begin{equation}\label{eq:spdm3}
    \mathbf{Y}(t) =\rho \mathbf{W}\mathbf{Y}(t)+\mathbf{X}(t)\boldsymbol{\beta}+\boldsymbol{\mu}+\boldsymbol{\epsilon}(t)
\end{equation}
\begin{equation}\label{eq:spdm4}
    \mathbf{Y}(t) = \rho \mathbf{W}\mathbf{X}(t) + \mathbf{X}(t)\boldsymbol{\beta}+\boldsymbol{\mu}+\boldsymbol{\epsilon}(t)
\end{equation}
\begin{equation}\label{eq:spdm5}
    \mathbf{Y}(t) = \mathbf{X}(t)\boldsymbol{\beta}+\boldsymbol{\mu}+\boldsymbol{\phi}(t)
\end{equation}
Spatial panel data models extend simple pooled linear regression with spatial specific effects and spatial interaction effects, but without considering the temporal autocorrelation effect (dependency at nearby time steps).

\emph{Spatiotemporal autoregressive model (STAR)}: Spatiotemporal autoregressive model (STAR) extends spatial autoregressive model (SAR) by incorporating temporal autoregression. Specifically, the SAR model can be generally written as $(\mathbf{I}-\rho \mathbf{W})\mathbf{Y}=\mathbf{X}\boldsymbol{\beta}+\boldsymbol{\epsilon}$. The term $\mathbf{I}-\rho \mathbf{W}$ helps to remove spatial autoregressive effect from the response vector $\mathbf{Y}$ so that residual errors $\boldsymbol{\epsilon}$ are i.i.d. Spatiotemporal autoregressive model generalizes $\mathbf{X}$ and $\mathbf{Y}$ from spatial data to spatiotemporal data (stacking all samples into different rows), and also extends $\mathbf{I}-\rho \mathbf{W}$ to incorporate temporal autoregressive effect $\mathbf{I}-\rho_S \mathbf{W}_S-\rho_T \mathbf{W}_T$ or spatiotemporal autoregressive effect $(\mathbf{I}-\rho_S \mathbf{W}_S)(\mathbf{I}-\rho_T \mathbf{W}_T)$, where $\mathbf{W}_S$ and $\mathbf{W}_T$ model spatial neighbors and temporal neighbors respectively.

\emph{Spatiotemporal Kriging}: Spatiotemporal Kriging is an extension of Kriging for spatiotemporal interpolation. It assumes that $n$ observations in continuous space and time $\{y(\mathbf{s_i},t_i)|i\in \mathbb{N}, 1\leq i\leq n\}$ follow a multi-variate Gaussian distribution $N(\boldsymbol{\mu},\boldsymbol{\Sigma})$. The mean vector $\boldsymbol{\mu}$ can be modeled as a linear function of sample covariates and coefficients, and the covariance matrix $\boldsymbol{\Sigma}$ can be determined based on \emph{spatiotemporal covariogram}. Assuming both spatial and temporal stationarity, covariance between two variables are location and time invariant, as in Equation~\ref{eq:stcovariogram}. The function 
\begin{equation}\label{eq:stcovariogram}
    Cov(y(\mathbf{s},t),y(\mathbf{s+h},t+r))\equiv C(\mathbf{h},r)
\end{equation} 
$C(\mathbf{h},r)$ is called spatiotemporal covariogram. Similar to Kriging, once spatiotemporal covariogram function is empirically estimated, we can derive the covariance matrix of joint Gaussian distribution based on sample location and time differences. Unknown observation at a new location and time $y(\mathbf{s_0},t_0)$ can be estimated through conditional expectation as shown in Equation~\ref{eq:stkriging}.
\begin{equation}\label{eq:stkriging}
    E(y(\mathbf{s_0},t_0)|y(\mathbf{s_i},t_i),1\leq i\leq n)
\end{equation}

\subsubsection{Temporal Non-stationarity}
Temporal non-stationarity means that sample distribution can be varying over time. This poses a challenge in that we cannot fit a same prediction model for all time steps. Addressing the challenge often requires the design of spatiotemporal dynamic models.

\emph{Spatiotemporal dynamic models}: Spatiotemporal dynamic models consider spatiotemporal observation data as a sequence of temporal snapshots,  $\{\mathbf{Y}(t)|t\in\mathbb{N},1\leq t \leq T\}$, where $\mathbf{Y}(t)=[y(\mathbf{s_1},t),...,y(\mathbf{s_n},t)]^T$, and $n$ is the number of spatial locations~\cite{cressie2015statistics2}. Moreover, observations depend on a corresponding sequence of temporal snapshots in the form of hidden variables (hidden process) $\{\mathbf{Z}(t)|t\in\mathbb{N},1\leq t \leq T\}$. Temporal autocorrelation (dependency) is usually modeled via Markov property on the hidden process. This is formally written in Equation~\ref{eq:stdynamic1} and Equation~\ref{eq:stdynamic2}, where $\boldsymbol{\epsilon}(t)$ and $\mathbf{e}(t)$ are noise, $f_t$ is a function that captures dependency 
of observations on hidden variables at a specific time step, and $g_t$ is a function that captures transitions of hidden variables over time. In the most simple form, $f_t$ and $g_t$ are linear, and $\boldsymbol{\epsilon}(t)$ and $\mathbf{e}(t)$ are Gaussian. There are other more complicated cases to incorporate spatial heterogeneity and temporal dynamics~\cite{stroud2001dynamic}. The hidden process variables can be configured based on physics or theories from an application domain, making spatiotemporal dynamic models useful tools in many fields such as meteorology and climate science.
\begin{equation}\label{eq:stdynamic1}
    \mathbf{Y}(t)=f_t(\mathbf{Z}(t))+\boldsymbol{\epsilon}(t)
\end{equation}
\begin{equation}\label{eq:stdynamic2}
    \mathbf{Z}(t)=g_t(\mathbf{Z}(t-1))+\mathbf{e}(t)
\end{equation}

\section{Future Research Opportunities}\label{sec:future}
This section summarizes future research opportunities. Most existing spatial prediction methods focus on the challenge of spatial autocorrelation. A few methods address spatial heterogeneity in the aspect of non-stationarity. Challenges of heterogeneous spatial data with anisotropic spatial dependency, multi-scale and multi-modality are largely underexplored. Moreover, the emergence of deep learning and spatial big data also represent new frontiers of spatial prediction research.

\subsection{Prediction for Heterogeneous Spatial Data}
Existing prediction methods addressing spatial heterogeneity focus on non-stationarity, assuming that training and test samples are within the same spatial framework, and that sample distribution is isotropic. However, in real world, spatial dependency can be anisotropic and test samples can be beyond the training area. Thus, novel spatial prediction methods need to be developed.

\emph{Prediction with anisotropic spatial dependency}: In real world spatial data, spatial dependency across sample locations can be anistropic, instead of being uniform in all directions in the Euclidean space. For example, dependency between observations on spatial networks (e.g., pollutants in river networks or traffic accidents on road networks) often follow network topology (e.g., river flow directions, traffic directions). As another instance of example, flood water locations (pixels) in an earth observation imagery implicitly follow terrains and topography due to gravity (i.e., water flows to a lower elevation). Anisotropic prediction on spatial networks poses unique challenges due to directional spatial dependency, and expensive network distance computation. Recently, several spatial statistical methods have been generalized from Euclidean space to spatial network space, e.g., network spatial autocorrelation, network kernel density, and network Kriging~\cite{okabe2012spatial}. However, little research has been done on prediction methods that incorporate anisotropic spatial dependency (i.e., spatial dependency along certain directions). Existing Markov random field methods only reflect undirected spatial dependency, and thus cannot be easily applied. Bayesian networks have been used to model directed dependency, but it is unscalable to a large number of nodes (locations). Recently, a hidden Markov tree model has been proposed to capture anisotropic spatial dependency by a reverse tree structure in the hidden class layer, but is studied in the context of flood mapping applications~\cite{jiangkdd2018,jiang2019hidden,jiang2019geographical,sainju2020hidden}. Addressing the challenge requires innovations in model structure, regularizers in objective functions, as well as effective and efficient learning algorithms.

\emph{Model transfer across heterogeneous spatial regions}: Due to the effect of spatial heterogeneity, prediction models learned from one region may not perform well in another. This is an issue since in many spatial prediction problems, test samples may not lie within the same spatial domain as training samples. In machine learning, similar issues have been studied through transfer learning~\cite{pan2010survey} and domain adaptation~\cite{daume2006domain}, but their corresponding methods for spatial prediction are largely under-explored. Addressing the challenge may require assumptions on the relationships or structure of spatial sample distributions between one region and another, as well as the fusion of auxiliary data to provide common spatial (or spatiotemporal) contexts.

\subsection{Data Fusion of Multi-scale Spatial Data from Different Sources}
Data fusion is the process of combining data from multiple sources to improve inference. Existing research on data fusion include multi-sensor fusion from signal processing perspective~\cite{hall1997introduction,waltz2001principle,khaleghi2013multisensor} and data integration from data management perspective~\cite{bleiholder2009data,dong2009data}. For spatial prediction, we are more interested in spatial data fusion that can improve predictive performance. A recent survey summarizes techniques to fuse cross-domain data for big data analytics~\cite{zheng2015methodologies}. Methods are categorized into stage-based, feature-level-based, and semantic meaning-based. 

\emph{Data fusion for multi-resolution earth imagery classification}: Earth observation imagery from different satellite and airborne platforms have different spatial, temporal, and spectral resolutions and coverage. Moreover, imagery from each single source is imperfect with noise, cloud, and obstacles. Spatial prediction methods that can utilize a diverse portfolio of earth imagery with multiple resolutions are of great practical value. Potential research directions include multi-view learning and multi-instance learning~\cite{karpatne2016monitoring}.

\emph{Data fusion for multi-modal spatial data}: In many spatial prediction applications, spatial data comes in different representations (e.g., points, line-strings, polygons, and raster imagery) and modalities (e.g., geo-social media, geotagged imagery and videos). For example, in precision agriculture, hyperspectral imagery often has high spatial details and complete spatial coverage, ground soil samples are only taken at several point locations, and crop yield are recorded at per-plot level. The goal is to predict crop yield in an early growing phase to optimize fertilizer allocation. Utilize such multi-modal data in spatial prediction requires data fusion and uncertainty quantification.

\subsection{Deep Learning Methods for Spatial Prediction}
Deep learning is a set of machine learning algorithms that use a multi-layer graph structure to extract a hierarchy of features at different levels~\cite{goodfellow2016deep}. High-level features in a top layer is built upon low-level features in lower layers. Deep learning models (e.g., deep convolutional neural network, deep recurrent neural network) have been shown successful in computer vision and natural language processing tasks. In the last couple of years, deep learning has been applied to spatial prediction problems, particularly on remote sensing imagery. Two recent surveys~\cite{zhang2016deep,zhu2017deep} summarize progress of utilizing deep learning techniques in classifying hyperspectral imagery for land cover mapping, radar imagery for target recognition, as well as  high-resolution aerial imagery for scene classification and object detection. Based on the types of inputs and outputs, methods can be categorized into per-pixel classification and per-image classification. In \emph{per-pixel} classification, the input network layer consists of spectral and spatial features extracted from spectral band values within the neighborhood of a pixel, and the output layer consists of thematic class categories for that pixel (e.g., land cover types). Based on a sliding window method, all pixels in the image can be classified. Recently, there are works that predict the classes of all pixels in one network architecture end-to-end, such as U-Net~\cite{ronneberger2015u}. In \emph{per-image} classification, the input layer consists of all pixels of an image, and the output layer consists of class categories of the entire image (e.g., scenes or object types). This research area is still growing rapidly with open challenges to address.

\emph{Limited ground truth class labels}: Large amount of training data is one important factor for the success of deep learning methods. Unfortunately, in remote sensing applications, collecting ground truth is both expensive and time consuming, as discussed in Section~\ref{subsec:limited}. There are several potential directions to address the challenge. In some problems such as high-resolution aerial imagery classification, we can leverage existing well-known datasets with similar data types (e.g., ImageNet dataset~\cite{deng2009imagenet}, IARPA functional map of the world challenge dataset~\cite{iarpa}, UC Merced land use dataset~\cite{yang2010bag}) to train a deep model or adapt well-trained deep models to a new application. For other problems, however, existing datasets and models may not be readily useful. In this case, collecting ground truth labels by well-trained experts at a large scale is infeasible. Several promising directions include utilizing crowd-sourcing from volunteered geographic information (e.g., Amazon Mechanical Turk, Tomnod.com by DigitalGlobe) and geotagged social media (e.g., tweets, Facebook posts), as well as leveraging physics-based modeling and simulation.

\emph{Enhancing interpretability}: One major limitation of deep learning is the lack of interpretability. This may not be a major concern in business applications, but in scientific fields such as climate science and hydrological science, interpretability is critical for the theoretical development of the field. One potential direction is to incorporate physical theories and constraints in model design and architecture, as generally discussed in theory-guided data science~\cite{karpatne2017theory}.

\subsection{Spatial Big Data Prediction}
Spatial big data (SBD)~\cite{shekhar2012spatial} refers to geo-referenced data whose volume, velocity, and variety exceed the capability of traditional spatial computational platforms. Examples of spatial big data include earth observation imagery (NASA collects petabytes of imagery each year~\cite{vatsavai2012spatiotemporal}), GPS trajectories, temporally detailed road networks, and cellphone check-in histories. Making predictions on spatial big data provides unique opportunities for large scale scientific studies such as national water forecasting and global land cover change analysis, but is also technically challenging due to the large data volume. 
In recent years, spatial big data techniques have been developed, including HadoopGIS~\cite{planthaber2012earthdb} and SpatialHadoop~\cite{eldawy2015spatialhadoop}, GeoSpark~\cite{yu2015geospark}, GPU-based algorithms~\cite{prasad2015vision,puri2013efficient,zhang2012speeding}, distributed database systems EarthDB~\cite{planthaber2012earthdb}, as well as Google Earth Engine~\cite{googleearthengine}. Currently, these existing platforms currently mostly focus on basic spatial operations (e.g., spatial joins), or traditional non-spatial prediction algorithms.  Thus, future research is needed to develop parallel spatial prediction algorithms on spatial big data platforms. 

\emph{Parallel spatial prediction algorithms}: Algorithm design and platform selection are determined by the computational structure of spatial prediction algorithms. Common computational structure includes filter-and-refine~\cite{shekhar2003spatial}, divid-and-conquer, matrix operations, and iterations (iteration is common in Expectation and Maximization, Newton Raphson, Markov Chain Monte Carlo simulation). Thus, GPUs and Spark platforms are potentially appropriate platforms.

\section{Conclusion}\label{sec:con}
This survey provides a systematic overview on spatial prediction techniques. Spatial prediction is of great importance in various application areas, such as earth science, urban informatics, social media analytics, and public health, but is technically challenging due to the unique characteristics of spatial data. We provide a taxonomy of spatial prediction methods based on the challenge they address and discuss several spatiotemporal extensions. We also identify future research opportunities.


%

\ifCLASSOPTIONcompsoc
  \section*{Acknowledgments}
\else
  \section*{Acknowledgment}
\fi
This material is based upon work supported by the NSF under Grant No. IIS-1850546, IIS-2008973, CNS-1951974, the National Oceanic and Atmospheric Administration (NOAA), and the University Corporation for Atmospheric Research (UCAR).

\ifCLASSOPTIONcaptionsoff
  \newpage
\fi



\bibliographystyle{IEEEtran}
\bibliography{SP_TKDE_Survey}

\begin{thebibliography}{100}
\providecommand{\url}[1]{#1}
\csname url@samestyle\endcsname
\providecommand{\newblock}{\relax}
\providecommand{\bibinfo}[2]{#2}
\providecommand{\BIBentrySTDinterwordspacing}{\spaceskip=0pt\relax}
\providecommand{\BIBentryALTinterwordstretchfactor}{4}
\providecommand{\BIBentryALTinterwordspacing}{\spaceskip=\fontdimen2\font plus
\BIBentryALTinterwordstretchfactor\fontdimen3\font minus
  \fontdimen4\font\relax}
\providecommand{\BIBforeignlanguage}[2]{{%
\expandafter\ifx\csname l@#1\endcsname\relax
\typeout{** WARNING: IEEEtran.bst: No hyphenation pattern has been}%
\typeout{** loaded for the language `#1'. Using the pattern for}%
\typeout{** the default language instead.}%
\else
\language=\csname l@#1\endcsname
\fi
#2}}
\providecommand{\BIBdecl}{\relax}
\BIBdecl

\bibitem{krige1951statistical}
D.~G. Krige, ``A statistical approach to some basic mine valuation problems on
  the witwatersrand,'' \emph{Journal of the Southern African Institute of
  Mining and Metallurgy}, vol.~52, no.~6, pp. 119--139, 1951.

\bibitem{shekhar2011identifying}
S.~Shekhar, M.~R. Evans, J.~M. Kang, and P.~Mohan, ``Identifying patterns in
  spatial information: A survey of methods,'' \emph{Wiley Interdisciplinary
  Reviews: Data Mining and Knowledge Discovery}, vol.~1, no.~3, pp. 193--214,
  2011.

\bibitem{jiang2017spatial}
Z.~Jiang and S.~Shekhar, \emph{Spatial Big Data Science: Classification
  Techniques for Earth Observation Imagery}.\hskip 1em plus 0.5em minus
  0.4em\relax Springer, 2017.

\bibitem{hansen2013high}
M.~C. Hansen, P.~V. Potapov, R.~Moore, M.~Hancher, S.~Turubanova, A.~Tyukavina,
  D.~Thau, S.~Stehman, S.~Goetz, T.~Loveland \emph{et~al.}, ``High-resolution
  global maps of 21st-century forest cover change,'' \emph{science}, vol. 342,
  no. 6160, pp. 850--853, 2013.

\bibitem{pekel2016high}
J.-F. Pekel, A.~Cottam, N.~Gorelick, and A.~S. Belward, ``High-resolution
  mapping of global surface water and its long-term changes,'' \emph{Nature},
  2016.

\bibitem{brivio2002integration}
P.~Brivio, R.~Colombo, M.~Maggi, and R.~Tomasoni, ``Integration of remote
  sensing data and gis for accurate mapping of flooded areas,''
  \emph{International Journal of Remote Sensing}, vol.~23, no.~3, pp. 429--441,
  2002.

\bibitem{jiangkdd2018}
M.~Xie, Z.~Jiang, and A.~M. Sainju, ``Geographical hidden markov tree for flood
  extent mapping,'' in \emph{Proceedings of the 24th {ACM} {SIGKDD}
  International Conference on Knowledge Discovery and Data Mining, London, UK,
  August 19-23, 2018}, 2018, pp. xx--xx.

\bibitem{moran1997opportunities}
M.~S. Moran, Y.~Inoue, and E.~Barnes, ``Opportunities and limitations for
  image-based remote sensing in precision crop management,'' \emph{Remote
  sensing of Environment}, vol.~61, no.~3, pp. 319--346, 1997.

\bibitem{austin2002spatial}
M.~Austin, ``Spatial prediction of species distribution: an interface between
  ecological theory and statistical modelling,'' \emph{Ecological modelling},
  vol. 157, no.~2, pp. 101--118, 2002.

\bibitem{elith2009species}
J.~Elith and J.~R. Leathwick, ``Species distribution models: ecological
  explanation and prediction across space and time,'' \emph{Annual review of
  ecology, evolution, and systematics}, vol.~40, pp. 677--697, 2009.

\bibitem{chang2001near}
C.-W. Chang, D.~A. Laird, M.~J. Mausbach, and C.~R. Hurburgh, ``Near-infrared
  reflectance spectroscopy--principal components regression analyses of soil
  properties,'' \emph{Soil Science Society of America Journal}, vol.~65, no.~2,
  pp. 480--490, 2001.

\bibitem{hengl2004generic}
T.~Hengl, G.~B. Heuvelink, and A.~Stein, ``A generic framework for spatial
  prediction of soil variables based on regression-kriging,'' \emph{Geoderma},
  vol. 120, no.~1, pp. 75--93, 2004.

\bibitem{Meng2017}
\BIBentryALTinterwordspacing
C.~Meng, X.~Yi, L.~Su, J.~Gao, and Y.~Zheng, ``City-wide traffic volume
  inference with loop detector data and taxi trajectories,'' in
  \emph{Proceedings of the 25th ACM SIGSPATIAL International Conference on
  Advances in Geographic Information Systems}, ser. SIGSPATIAL'17.\hskip 1em
  plus 0.5em minus 0.4em\relax New York, NY, USA: ACM, 2017, pp. 1:1--1:10.
  [Online]. Available: \url{http://doi.acm.org/10.1145/3139958.3139984}
\BIBentrySTDinterwordspacing

\bibitem{demandPredAAAI18}
H.~Yao, F.~Wu, J.~Ke, X.~Tang, Y.~Jia, S.~Lu, S.~Gong, J.~Ye, and Z.~Li, ``Deep
  multi-view spatial-temporal network for taxi demand prediction,'' in
  \emph{Proceedings of the International Conference on Artificial
  Intelligence}.\hskip 1em plus 0.5em minus 0.4em\relax AAAI Press, 2018, pp.
  0--10.

\bibitem{sakaki2010earthquake}
T.~Sakaki, M.~Okazaki, and Y.~Matsuo, ``Earthquake shakes twitter users:
  real-time event detection by social sensors,'' in \emph{Proceedings of the
  19th international conference on World wide web}.\hskip 1em plus 0.5em minus
  0.4em\relax ACM, 2010, pp. 851--860.

\bibitem{zhang2017triovecevent}
C.~Zhang, L.~Liu, D.~Lei, Q.~Yuan, H.~Zhuang, T.~Hanratty, and J.~Han,
  ``Triovecevent: Embedding-based online local event detection in geo-tagged
  tweet streams,'' in \emph{Proceedings of the 23rd ACM SIGKDD International
  Conference on Knowledge Discovery and Data Mining}.\hskip 1em plus 0.5em
  minus 0.4em\relax ACM, 2017, pp. 595--604.

\bibitem{DBLP:conf/kdd/ZhaoSYCLR15}
\BIBentryALTinterwordspacing
L.~Zhao, Q.~Sun, J.~Ye, F.~Chen, C.~Lu, and N.~Ramakrishnan, ``Multi-task
  learning for spatio-temporal event forecasting,'' in \emph{Proceedings of the
  21th {ACM} {SIGKDD} International Conference on Knowledge Discovery and Data
  Mining, Sydney, NSW, Australia, August 10-13, 2015}, 2015, pp. 1503--1512.
  [Online]. Available: \url{http://doi.acm.org/10.1145/2783258.2783377}
\BIBentrySTDinterwordspacing

\bibitem{majid2013context}
A.~Majid, L.~Chen, G.~Chen, H.~T. Mirza, I.~Hussain, and J.~Woodward, ``A
  context-aware personalized travel recommendation system based on geotagged
  social media data mining,'' \emph{International Journal of Geographical
  Information Science}, vol.~27, no.~4, pp. 662--684, 2013.

\bibitem{best2005comparison}
N.~Best, S.~Richardson, and A.~Thomson, ``A comparison of bayesian spatial
  models for disease mapping,'' \emph{Statistical methods in medical research},
  vol.~14, no.~1, pp. 35--59, 2005.

\bibitem{rappaport2010environment}
S.~M. Rappaport and M.~T. Smith, ``Environment and disease risks,''
  \emph{Science}, vol. 330, no. 6003, pp. 460--461, 2010.

\bibitem{lee2016mind}
E.~C. Lee, J.~M. Asher, S.~Goldlust, J.~D. Kraemer, A.~B. Lawson, and
  S.~Bansal, ``Mind the scales: Harnessing spatial big data for infectious
  disease surveillance and inference,'' \emph{The Journal of infectious
  diseases}, vol. 214, no. suppl\_4, pp. S409--S413, 2016.

\bibitem{worboys2004gis}
M.~F. Worboys and M.~Duckham, \emph{GIS: a computing perspective}.\hskip 1em
  plus 0.5em minus 0.4em\relax CRC press, 2004.

\bibitem{tobler1970computer}
W.~R. Tobler, ``A computer movie simulating urban growth in the detroit
  region,'' \emph{Economic geography}, vol.~46, no. sup1, pp. 234--240, 1970.

\bibitem{jiang2015focal}
Z.~Jiang, S.~Shekhar, X.~Zhou, J.~Knight, and J.~Corcoran, ``Focal-test-based
  spatial decision tree learning,'' \emph{IEEE Transactions on Knowledge and
  Data Engineering}, vol.~27, no.~6, pp. 1547--1559, 2015.

\bibitem{ess2001culture}
C.~Ess and F.~Sudweeks, \emph{Culture, technology, communication: Towards an
  intercultural global village}.\hskip 1em plus 0.5em minus 0.4em\relax Suny
  Press, 2001.

\bibitem{congalton1991review}
R.~G. Congalton, ``A review of assessing the accuracy of classifications of
  remotely sensed data,'' \emph{Remote sensing of environment}, vol.~37, no.~1,
  pp. 35--46, 1991.

\bibitem{wilby2004guidelines}
R.~L. Wilby, S.~Charles, E.~Zorita, B.~Timbal, P.~Whetton, and L.~Mearns,
  ``Guidelines for use of climate scenarios developed from statistical
  downscaling methods,'' \emph{Supporting material of the Intergovernmental
  Panel on Climate Change, available from the DDC of IPCC TGCIA}, vol.~27,
  2004.

\bibitem{ester1997spatial}
M.~Ester, H.-P. Kriegel, and J.~Sander, ``Spatial data mining: A database
  approach,'' in \emph{International Symposium on Spatial Databases}.\hskip 1em
  plus 0.5em minus 0.4em\relax Springer, 1997, pp. 47--66.

\bibitem{koperski1996spatial}
K.~Koperski, J.~Adhikary, and J.~Han, ``Spatial data mining: progress and
  challenges survey paper,'' in \emph{Proc. ACM SIGMOD Workshop on Research
  Issues on Data Mining and Knowledge Discovery, Montreal, Canada}.\hskip 1em
  plus 0.5em minus 0.4em\relax Citeseer, 1996, pp. 1--10.

\bibitem{miller2009geographic}
H.~J. Miller and J.~Han, \emph{Geographic data mining and knowledge
  discovery}.\hskip 1em plus 0.5em minus 0.4em\relax CRC Press, 2009.

\bibitem{shekhar2003trends}
S.~Shekhar, P.~Zhang, Y.~Huang, and R.~Vatsavai, ``Trends in spatial data
  mining. as a chapter in data mining: Next generation challenges and future
  directions, h. kargupta, a. joshi, k. sivakumar, and y. yesha,'' 2003.

\bibitem{shekhar2015spatiotemporal}
S.~Shekhar, Z.~Jiang, R.~Y. Ali, E.~Eftelioglu, X.~Tang, V.~Gunturi, and
  X.~Zhou, ``Spatiotemporal data mining: A computational perspective,''
  \emph{ISPRS International Journal of Geo-Information}, vol.~4, no.~4, pp.
  2306--2338, 2015.

\bibitem{atluri2017spatio}
G.~Atluri, A.~Karpatne, and V.~Kumar, ``Spatio-temporal data mining: A survey
  of problems and methods,'' \emph{arXiv preprint arXiv:1711.04710}, 2017.

\bibitem{schabenberger2005statistical}
O.~Schabenberger and C.~Gotway, \emph{Statistical methods for spatial data
  analysis}.\hskip 1em plus 0.5em minus 0.4em\relax CRC Press, 2005, vol.~64.

\bibitem{banerjee2014hierarchical}
S.~Banerjee, B.~P. Carlin, and A.~E. Gelfand, \emph{Hierarchical modeling and
  analysis for spatial data}.\hskip 1em plus 0.5em minus 0.4em\relax Crc Press,
  2014.

\bibitem{cressie2015statistics}
N.~Cressie, \emph{Statistics for spatial data}.\hskip 1em plus 0.5em minus
  0.4em\relax John Wiley \& Sons, 2015.

\bibitem{mcgovern_spatiotemporal_2008}
A.~McGovern, N.~C. Hiers, M.~W. Collier, D.~J.~G. II, and R.~A. Brown,
  ``Spatiotemporal relational probability trees: An introduction,'' in
  \emph{{ICDM}}, 2008, pp. 935--940.

\bibitem{mcgovern_enhanced_2013}
A.~McGovern, N.~Troutman, R.~A. Brown, J.~K. Williams, and J.~Abernethy,
  ``Enhanced spatiotemporal relational probability trees and forests,''
  \emph{Data Min. Knowl. Discov.}, vol.~26, no.~2, pp. 398--433, 2013.

\bibitem{frank_multi-relational_2009}
R.~Frank, M.~Ester, and A.~J. Knobbe, ``A multi-relational approach to spatial
  classification,'' in \emph{{KDD}}, 2009, pp. 309--318.

\bibitem{ding_discovery_2009}
W.~Ding, T.~F. Stepinski, and J.~Salazar, ``Discovery of geospatial
  discriminating patterns from remote sensing datasets,'' in \emph{{SDM}},
  2009, pp. 425--436.

\bibitem{lu2007survey}
D.~Lu and Q.~Weng, ``A survey of image classification methods and techniques
  for improving classification performance,'' \emph{International journal of
  Remote sensing}, vol.~28, no.~5, pp. 823--870, 2007.

\bibitem{chan2005salt}
R.~H. Chan, C.-W. Ho, and M.~Nikolova, ``Salt-and-pepper noise removal by
  median-type noise detectors and detail-preserving regularization,''
  \emph{Image Processing, IEEE Transactions on}, vol.~14, no.~10, pp.
  1479--1485, 2005.

\bibitem{brownrigg1984weighted}
D.~Brownrigg, ``The weighted median filter,'' \emph{Communications of the ACM},
  vol.~27, no.~8, pp. 807--818, 1984.

\bibitem{hwang1995adaptive}
H.~Hwang and R.~A. Haddad, ``Adaptive median filters: new algorithms and
  results,'' \emph{Image Processing, IEEE Transactions on}, vol.~4, no.~4, pp.
  499--502, 1995.

\bibitem{esakkirajan2011removal}
S.~Esakkirajan, T.~Veerakumar, A.~N. Subramanyam, and C.~PremChand, ``Removal
  of high density salt and pepper noise through modified decision based
  unsymmetric trimmed median filter,'' \emph{Signal Processing Letters, IEEE},
  vol.~18, no.~5, pp. 287--290, 2011.

\bibitem{puissant2005utility}
A.~Puissant, J.~Hirsch, and C.~Weber, ``The utility of texture analysis to
  improve per-pixel classification for high to very high spatial resolution
  imagery,'' \emph{International Journal of Remote Sensing}, vol.~26, no.~4,
  pp. 733--745, 2005.

\bibitem{jiang_focal-test-based_2013}
Z.~Jiang, S.~Shekhar, X.~Zhou, J.~Knight, and J.~Corcoran, ``Focal-test-based
  spatial decision tree learning: A summary of results,'' in \emph{{ICDM}},
  2013, pp. 320--329.

\bibitem{benediktsson2005classification}
J.~A. Benediktsson, J.~A. Palmason, and J.~R. Sveinsson, ``Classification of
  hyperspectral data from urban areas based on extended morphological
  profiles,'' \emph{IEEE Transactions on Geoscience and Remote Sensing},
  vol.~43, no.~3, pp. 480--491, 2005.

\bibitem{hay2008geographic}
G.~Hay and G.~Castilla, ``Geographic object-based image analysis (geobia): A
  new name for a new discipline,'' in \emph{Object-based image analysis}.\hskip
  1em plus 0.5em minus 0.4em\relax Springer, 2008, pp. 75--89.

\bibitem{tarabalka2009spectral}
Y.~Tarabalka, J.~A. Benediktsson, and J.~Chanussot, ``Spectral--spatial
  classification of hyperspectral imagery based on partitional clustering
  techniques,'' \emph{Geoscience and Remote Sensing, IEEE Transactions on},
  vol.~47, no.~8, pp. 2973--2987, 2009.

\bibitem{zheng2013u}
Y.~Zheng, F.~Liu, and H.-P. Hsieh, ``U-air: When urban air quality inference
  meets big data,'' in \emph{Proceedings of the 19th ACM SIGKDD international
  conference on Knowledge discovery and data mining}.\hskip 1em plus 0.5em
  minus 0.4em\relax ACM, 2013, pp. 1436--1444.

\bibitem{wu2015semantic}
F.~Wu, Z.~Li, W.-C. Lee, H.~Wang, and Z.~Huang, ``Semantic annotation of
  mobility data using social media,'' in \emph{Proceedings of the 24th
  International Conference on World Wide Web}.\hskip 1em plus 0.5em minus
  0.4em\relax International World Wide Web Conferences Steering Committee,
  2015, pp. 1253--1263.

\bibitem{wu2016did}
F.~Wu and Z.~Li, ``Where did you go: Personalized annotation of mobility
  records,'' in \emph{Proceedings of the 25th ACM International on Conference
  on Information and Knowledge Management}.\hskip 1em plus 0.5em minus
  0.4em\relax ACM, 2016, pp. 589--598.

\bibitem{zheng2015methodologies}
Y.~Zheng, ``Methodologies for cross-domain data fusion: An overview,''
  \emph{IEEE transactions on big data}, vol.~1, no.~1, pp. 16--34, 2015.

\bibitem{brook1964distinction}
D.~Brook, ``On the distinction between the conditional probability and the
  joint probability approaches in the specification of nearest-neighbour
  systems,'' \emph{Biometrika}, vol.~51, no. 3/4, pp. 481--483, 1964.

\bibitem{clifford1990markov}
P.~Clifford, ``Markov random fields in statistics,'' \emph{Disorder in physical
  systems: A volume in honour of John M. Hammersley}, pp. 19--32, 1990.

\bibitem{anse-88}
L.~Anselin, \emph{{Spatial Econometrics: methods and models}}.\hskip 1em plus
  0.5em minus 0.4em\relax Dordrecht, Netherlands: Kluwer, 1988.

\bibitem{viton2010notes}
P.~A. Viton, ``Notes on spatial econometric models,'' \emph{City and regional
  planning}, vol. 870, no.~03, pp. 9--10, 2010.

\bibitem{assunccao2009neighborhood}
R.~Assun{\c{c}}{\~a}o and E.~Krainski, ``Neighborhood dependence in bayesian
  spatial models,'' \emph{Biometrical Journal}, vol.~51, no.~5, pp. 851--869,
  2009.

\bibitem{li2009markov}
S.~Z. Li, \emph{Markov random field modeling in image analysis}.\hskip 1em plus
  0.5em minus 0.4em\relax Springer Science \& Business Media, 2009.

\bibitem{chawla_modeling_2001}
S.~Chawla, S.~Shekhar, W.~Wu, and U.~Ozesmi, ``Modeling spatial dependencies
  for mining geospatial data,'' in \emph{{SDM}}, 2001, pp. 1--17.

\bibitem{Shekhar-02}
S.~Shekhar, P.~R. Schrater, R.~R. Vatsavai, W.~Wu, and S.~Chawla, ``{Spatial
  Contextual Classification and Prediction Models for Mining Geospatial
  Data},'' \emph{IEEE Transactions on Multimedia}, vol.~4, no.~2, 2002.

\bibitem{boykov2001fast}
Y.~Boykov, O.~Veksler, and R.~Zabih, ``Fast approximate energy minimization via
  graph cuts,'' \emph{IEEE Transactions on pattern analysis and machine
  intelligence}, vol.~23, no.~11, pp. 1222--1239, 2001.

\bibitem{besag1986statistical}
J.~Besag, ``On the statistical analysis of dirty pictures,'' \emph{Journal of
  the Royal Statistical Society. Series B (Methodological)}, pp. 259--302,
  1986.

\bibitem{jackson2002adaptive}
Q.~Jackson and D.~A. Landgrebe, ``Adaptive bayesian contextual classification
  based on markov random fields,'' \emph{IEEE Transactions on Geoscience and
  Remote Sensing}, vol.~40, no.~11, pp. 2454--2463, 2002.

\bibitem{fu_drought_2012}
Q.~Fu, A.~Banerjee, S.~Liess, and P.~K. Snyder, ``Drought detection of the last
  century: An {MRF}-based approach,'' in \emph{{SDM}}, 2012, pp. 24--34.

\bibitem{lafferty2001conditional}
J.~Lafferty, A.~McCallum, F.~Pereira \emph{et~al.}, ``Conditional random
  fields: Probabilistic models for segmenting and labeling sequence data,'' in
  \emph{Proceedings of the eighteenth international conference on machine
  learning, ICML}, vol.~1, 2001, pp. 282--289.

\bibitem{lee_efficient_2006}
C.-H. Lee, R.~Greiner, and O.~R. Za{\"\i}ane, ``Efficient spatial
  classification using decoupled conditional random fields,'' in \emph{{PKDD}},
  2006, pp. 272--283.

\bibitem{kumar2003discriminative}
S.~Kumar and M.~Hebert, ``Discriminative fields for modeling spatial
  dependencies in natural images.'' in \emph{NIPS}, vol.~16, no. 2003, 2003,
  pp. 1531--1538.

\bibitem{lee_support_2005}
C.-H. Lee, R.~Greiner, and M.~W. Schmidt, ``Support vector random fields for
  spatial classification,'' in \emph{{PKDD}}, 2005, pp. 121--132.

\bibitem{zimmerman1999experimental}
D.~Zimmerman, C.~Pavlik, A.~Ruggles, and M.~P. Armstrong, ``An experimental
  comparison of ordinary and universal kriging and inverse distance
  weighting,'' \emph{Mathematical Geology}, vol.~31, no.~4, pp. 375--390, 1999.

\bibitem{DBLP:conf/kdd/KimYTM15}
\BIBentryALTinterwordspacing
T.~Kim, Y.~Yue, S.~L. Taylor, and I.~A. Matthews, ``A decision tree framework
  for spatiotemporal sequence prediction,'' in \emph{Proceedings of the 21th
  {ACM} {SIGKDD} International Conference on Knowledge Discovery and Data
  Mining, Sydney, NSW, Australia, August 10-13, 2015}, 2015, pp. 577--586.
  [Online]. Available: \url{http://doi.acm.org/10.1145/2783258.2783356}
\BIBentrySTDinterwordspacing

\bibitem{DBLP:conf/kdd/FuXGYZZ14}
\BIBentryALTinterwordspacing
Y.~Fu, H.~Xiong, Y.~Ge, Z.~Yao, Y.~Zheng, and Z.~Zhou, ``Exploiting geographic
  dependencies for real estate appraisal: a mutual perspective of ranking and
  clustering,'' in \emph{The 20th {ACM} {SIGKDD} International Conference on
  Knowledge Discovery and Data Mining, {KDD} '14, New York, NY, {USA} - August
  24 - 27, 2014}, 2014, pp. 1047--1056. [Online]. Available:
  \url{http://doi.acm.org/10.1145/2623330.2623675}
\BIBentrySTDinterwordspacing

\bibitem{DBLP:conf/sdm/ZhaoCLR15}
\BIBentryALTinterwordspacing
L.~Zhao, F.~Chen, C.~Lu, and N.~Ramakrishnan, ``Spatiotemporal event
  forecasting in social media,'' in \emph{Proceedings of the 2015 {SIAM}
  International Conference on Data Mining, Vancouver, BC, Canada, April 30 -
  May 2, 2015}, 2015, pp. 963--971. [Online]. Available:
  \url{http://dx.doi.org/10.1137/1.9781611974010.108}
\BIBentrySTDinterwordspacing

\bibitem{weinberger2007graph}
K.~Q. Weinberger, F.~Sha, Q.~Zhu, and L.~K. Saul, ``Graph laplacian
  regularization for large-scale semidefinite programming,'' \emph{Advances in
  neural information processing systems}, vol.~19, p. 1489, 2007.

\bibitem{subbian_climate_2013}
K.~Subbian and A.~Banerjee, ``Climate multi-model regression using spatial
  smoothing,'' in \emph{{SDM}}, 2013, pp. 324--332.

\bibitem{DBLP:conf/sdm/DataKKBK14}
A.~Karpatne, A.~Khandelwal, S.~Boriah, and V.~Kumar, ``Predictive learning in
  the presence of heterogeneity and limited training data,'' in
  \emph{Proceedings of the 2014 {SIAM} International Conference on Data Mining,
  Philadelphia, Pennsylvania, USA, April 24-26, 2014}, 2014, pp. 253--261.

\bibitem{stoeckel_svm_2005}
J.~Stoeckel and G.~Fung, ``{SVM} feature selection for classification of
  {SPECT} images of alzheimer's disease using spatial information,'' in
  \emph{{ICDM}}, 2005, pp. 410--417.

\bibitem{avriel2003nonlinear}
M.~Avriel, \emph{Nonlinear programming: analysis and methods}.\hskip 1em plus
  0.5em minus 0.4em\relax Courier Corporation, 2003.

\bibitem{hansen2000global}
M.~Hansen, R.~DeFries, J.~R. Townshend, and R.~Sohlberg, ``Global land cover
  classification at 1 km spatial resolution using a classification tree
  approach,'' \emph{International journal of remote sensing}, vol.~21, no. 6-7,
  pp. 1331--1364, 2000.

\bibitem{pal2003assessment}
M.~Pal and P.~M. Mather, ``An assessment of the effectiveness of decision tree
  methods for land cover classification,'' \emph{Remote sensing of
  environment}, vol.~86, no.~4, pp. 554--565, 2003.

\bibitem{jiang_learning_2012}
Z.~Jiang, S.~Shekhar, P.~Mohan, J.~Knight, and J.~Corcoran, ``Learning spatial
  decision tree for geographical classification: a summary of results,'' in
  \emph{{SIGSPATIAL}/{GIS}}, 2012, pp. 390--393.

\bibitem{li_spatial_2006}
X.~Li and C.~Claramunt, ``A spatial {Entropy-Based} decision tree for
  classification of geographical information,'' \emph{Transactions in {GIS}},
  vol.~10, no.~3, pp. 451--467, Blackwell Publishing Ltd, 2006.

\bibitem{stojanova2011global}
D.~Stojanova, M.~Ceci, A.~Appice, D.~Malerba, and S.~D{\v{z}}eroski, ``Global
  and local spatial autocorrelation in predictive clustering trees,'' in
  \emph{International Conference on Discovery Science}.\hskip 1em plus 0.5em
  minus 0.4em\relax Springer, 2011, pp. 307--322.

\bibitem{stojanova2012dealing}
D.~Stojanova, M.~Ceci, A.~Appice, D.~Malerba, and S.~D{{z}}eroski, ``Dealing
  with spatial autocorrelation when learning predictive clustering trees,''
  \emph{Ecological Informatics, Elsevier}, 2012.

\bibitem{fotheringham2002geographically}
A.~Fotheringham, C.~Brunsdon, and M.~Charlton, \emph{{Geographically weighted
  regression: the analysis of spatially varying relationships}}.\hskip 1em plus
  0.5em minus 0.4em\relax Wiley, 2002.

\bibitem{nakaya2005geographically}
T.~Nakaya, A.~S. Fotheringham, C.~Brunsdon, and M.~Charlton, ``Geographically
  weighted poisson regression for disease association mapping,''
  \emph{Statistics in medicine}, vol.~24, no.~17, pp. 2695--2717, 2005.

\bibitem{harris2011geographically}
P.~Harris, C.~Brunsdon, and M.~Charlton, ``Geographically weighted principal
  components analysis,'' \emph{International Journal of Geographical
  Information Science}, vol.~25, no.~10, pp. 1717--1736, 2011.

\bibitem{ren2016ensemble}
Y.~Ren, L.~Zhang, and P.~Suganthan, ``Ensemble classification and
  regression-recent developments, applications and future directions [review
  article],'' \emph{Computational Intelligence Magazine, IEEE}, vol.~11, no.~1,
  pp. 41--53, 2016.

\bibitem{zhou2012ensemble}
Z.-H. Zhou, \emph{Ensemble methods: foundations and algorithms}.\hskip 1em plus
  0.5em minus 0.4em\relax CRC Press, 2012.

\bibitem{dietterich2000ensemble}
T.~G. Dietterich, ``Ensemble methods in machine learning,'' in \emph{Multiple
  classifier systems}.\hskip 1em plus 0.5em minus 0.4em\relax Springer, 2000,
  pp. 1--15.

\bibitem{jacobs1991adaptive}
R.~A. Jacobs, M.~I. Jordan, S.~J. Nowlan, and G.~E. Hinton, ``Adaptive mixtures
  of local experts,'' \emph{Neural computation}, vol.~3, no.~1, pp. 79--87,
  1991.

\bibitem{jordan1994hierarchical}
M.~I. Jordan and R.~A. Jacobs, ``Hierarchical mixtures of experts and the em
  algorithm,'' \emph{Neural computation}, vol.~6, no.~2, pp. 181--214, 1994.

\bibitem{KarpatneK15}
A.~Karpatne and V.~Kumar, ``Adaptive heterogeneous ensemble learning using the
  context of test instances,'' in \emph{2015 {IEEE} International Conference on
  Data Mining, {ICDM} 2015, Atlantic City, NJ, USA, November 14-17, 2015},
  2015, pp. 787--792.

\bibitem{karpatne2015ensemble}
\BIBentryALTinterwordspacing
A.~Karpatne, A.~Khandelwal, and V.~Kumar, ``Ensemble learning methods for
  binary classification with multi-modality within the classes,'' in
  \emph{Proceedings of the {SIAM} International Conference on Data Mining,
  2015}.\hskip 1em plus 0.5em minus 0.4em\relax {SIAM}, 2015, pp. 730--738.
  [Online]. Available: \url{http://dx.doi.org/10.1137/1.9781611974010.82}
\BIBentrySTDinterwordspacing

\bibitem{vatsavai_hybrid_2011}
R.~R. Vatsavai and B.~L. Bhaduri, ``A hybrid classification scheme for mining
  multisource geospatial data,'' \emph{{GeoInformatica}}, vol.~15, no.~1, pp.
  29--47, 2011.

\bibitem{jiang_ensemble_2017}
Z.~Jiang, Y.~Li, S.~Shekhar, L.~Rampi, and J.~Knight, ``Spatial ensemble
  learning for heterogeneous geographic data with class ambiguity: A summary of
  results,'' in \emph{Proceedings of the 25th ACM SIGSPATIAL International
  Conference on Advances in Geographic Information Systems}, ser.
  SIGSPATIAL'17.\hskip 1em plus 0.5em minus 0.4em\relax ACM, 2017, pp.
  23:1--23:10.

\bibitem{radosavljevic_spatio-temporal_2008}
V.~Radosavljevic, S.~Vucetic, and Z.~Obradovic, ``Spatio-temporal partitioning
  for improving aerosol prediction accuracy,'' in \emph{{SDM}}, 2008, pp.
  609--620.

\bibitem{Caruana1997}
R.~Caruana, ``Multitask learning,'' \emph{Machine Learning}, vol.~28, no.~1,
  pp. 41--75, 1997.

\bibitem{gonccalves2015multi}
A.~R. Gon{\c{c}}alves, F.~J. Von~Zuben, and A.~Banerjee, ``Multi-label
  structure learning with ising model selection,'' in \emph{Proceedings of the
  24th International Conference on Artificial Intelligence}.\hskip 1em plus
  0.5em minus 0.4em\relax AAAI Press, 2015, pp. 3525--3531.

\bibitem{zhu2005semi}
X.~Zhu, ``Semi-supervised learning literature survey,'' 2005.

\bibitem{dempster1977maximum}
A.~P. Dempster, N.~M. Laird, and D.~B. Rubin, ``Maximum likelihood from
  incomplete data via the em algorithm,'' \emph{Journal of the royal
  statistical society. Series B (methodological)}, pp. 1--38, 1977.

\bibitem{DBLP:conf/icdm/VatsavaiSB08}
\BIBentryALTinterwordspacing
R.~R. Vatsavai, S.~Shekhar, and B.~L. Bhaduri, ``A semi-supervised learning
  algorithm for recognizing sub-classes,'' in \emph{Workshops Proceedings of
  the 8th {IEEE} International Conference on Data Mining {(ICDM} 2008),
  December 15-19, 2008, Pisa, Italy}, 2008, pp. 458--467. [Online]. Available:
  \url{http://dx.doi.org/10.1109/ICDMW.2008.129}
\BIBentrySTDinterwordspacing

\bibitem{DBLP:conf/igarss/VatsavaiBSB08}
R.~R. Vatsavai, B.~L. Badhuri, S.~Shekhar, and T.~E. Burk, ``Multisource data
  classification using a hybrid semi-supervised learning scheme,'' in
  \emph{{IEEE} International Geoscience {\&} Remote Sensing Symposium, {IGARSS}
  2008, July 8-11, 2008, Boston, Massachusetts, USA, Proceedings}, 2008, pp.
  1016--1019.

\bibitem{DBLP:journals/paapp/VatsavaiSB07}
R.~R. Vatsavai, S.~Shekhar, and T.~E. Burk, ``An efficient spatial
  semi-supervised learning algorithm,'' \emph{{IJPEDS}}, vol.~22, no.~6, pp.
  427--437, 2007.

\bibitem{camps2007semi}
G.~Camps-Valls, T.~V.~B. Marsheva, and D.~Zhou, ``Semi-supervised graph-based
  hyperspectral image classification,'' \emph{IEEE Transactions on Geoscience
  and Remote Sensing}, vol.~45, no.~10, pp. 3044--3054, 2007.

\bibitem{gomez2008semisupervised}
L.~G{\'o}mez-Chova, G.~Camps-Valls, J.~Munoz-Mari, and J.~Calpe,
  ``Semisupervised image classification with laplacian support vector
  machines,'' \emph{IEEE Geoscience and Remote Sensing Letters}, vol.~5, no.~3,
  pp. 336--340, 2008.

\bibitem{dopido2013semisupervised}
I.~D{\'o}pido, J.~Li, P.~R. Marpu, A.~Plaza, J.~M.~B. Dias, and J.~A.
  Benediktsson, ``Semisupervised self-learning for hyperspectral image
  classification,'' \emph{IEEE Transactions on Geoscience and Remote Sensing},
  vol.~51, no.~7, pp. 4032--4044, 2013.

\bibitem{tan2015novel}
K.~Tan, J.~Hu, J.~Li, and P.~Du, ``A novel semi-supervised hyperspectral image
  classification approach based on spatial neighborhood information and
  classifier combination,'' \emph{ISPRS Journal of Photogrammetry and Remote
  Sensing}, vol. 105, pp. 19--29, 2015.

\bibitem{hong2015spatial}
Y.~Hong and W.~Zhu, ``Spatial co-training for semi-supervised image
  classification,'' \emph{Pattern Recognition Letters}, vol.~63, pp. 59--65,
  2015.

\bibitem{zhang2014modified}
X.~Zhang, Q.~Song, R.~Liu, W.~Wang, and L.~Jiao, ``Modified co-training with
  spectral and spatial views for semisupervised hyperspectral image
  classification,'' \emph{IEEE Journal of Selected Topics in Applied Earth
  Observations and Remote Sensing}, vol.~7, no.~6, pp. 2044--2055, 2014.

\bibitem{landsat}
{United States Geological Survey}, ``Landsat missions,''
  \url{https://landsat.usgs.gov/}.

\bibitem{modis}
NASA, ``Modis moderate resolution imaging spectroradiometer,''
  \url{https://modis.gsfc.nasa.gov/}.

\bibitem{settles2010active}
B.~Settles, ``Active learning literature survey,'' \emph{University of
  Wisconsin, Madison}, vol.~52, no. 55-66, p.~11, 2010.

\bibitem{tuia2011survey}
D.~Tuia, M.~Volpi, L.~Copa, M.~Kanevski, and J.~Munoz-Mari, ``A survey of
  active learning algorithms for supervised remote sensing image
  classification,'' \emph{IEEE Journal of Selected Topics in Signal
  Processing}, vol.~5, no.~3, pp. 606--617, 2011.

\bibitem{stumpf2014active}
A.~Stumpf, N.~Lachiche, J.-P. Malet, N.~Kerle, and A.~Puissant, ``Active
  learning in the spatial domain for remote sensing image classification,''
  \emph{IEEE Transactions on Geoscience and Remote Sensing}, vol.~52, no.~5,
  pp. 2492--2507, 2014.

\bibitem{liu_spatially_2009}
A.~Liu, G.~Jun, and J.~Ghosh, ``Spatially cost-sensitive active learning,'' in
  \emph{{SDM}}, 2009, pp. 814--825.

\bibitem{wong2009modifiable}
D.~Wong, ``The modifiable areal unit problem (maup),'' \emph{The SAGE handbook
  of spatial analysis}, pp. 105--123, 2009.

\bibitem{DBLP:conf/kdd/ZhaoYCLR16}
L.~Zhao, J.~Ye, F.~Chen, C.~Lu, and N.~Ramakrishnan, ``Hierarchical incomplete
  multi-source feature learning for spatiotemporal event forecasting,'' in
  \emph{Proceedings of the 22nd {ACM} {SIGKDD} International Conference on
  Knowledge Discovery and Data Mining, San Francisco, CA, USA, August 13-17,
  2016}, 2016, pp. 2085--2094.

\bibitem{bandyopadhyay2009bayesian}
D.~Bandyopadhyay, B.~J. Reich, and E.~H. Slate, ``Bayesian modeling of
  multivariate spatial binary data with applications to dental caries,''
  \emph{Statistics in medicine}, vol.~28, no.~28, pp. 3492--3508, 2009.

\bibitem{yu2016hierarchical}
C.-H. Yu, W.~Ding, M.~Morabito, and P.~Chen, ``Hierarchical spatio-temporal
  pattern discovery and predictive modeling,'' \emph{IEEE Transactions on
  Knowledge and Data Engineering}, vol.~28, no.~4, pp. 979--993, 2016.

\bibitem{mennis2005cubic}
J.~Mennis, R.~Viger, and C.~D. Tomlin, ``Cubic map algebra functions for
  spatio-temporal analysis,'' \emph{Cartography and Geographic Information
  Science}, vol.~32, no.~1, pp. 17--32, 2005.

\bibitem{fischer2009handbook}
M.~M. Fischer and A.~Getis, \emph{Handbook of applied spatial analysis:
  software tools, methods and applications}.\hskip 1em plus 0.5em minus
  0.4em\relax Springer Science \& Business Media, 2009.

\bibitem{cressie2015statistics2}
N.~Cressie and C.~K. Wikle, \emph{Statistics for spatio-temporal data}.\hskip
  1em plus 0.5em minus 0.4em\relax John Wiley \& Sons, 2015.

\bibitem{stroud2001dynamic}
J.~R. Stroud, P.~M{\"u}ller, and B.~Sans{\'o}, ``Dynamic models for
  spatiotemporal data,'' \emph{Journal of the Royal Statistical Society: Series
  B (Statistical Methodology)}, vol.~63, no.~4, pp. 673--689, 2001.

\bibitem{okabe2012spatial}
A.~Okabe and K.~Sugihara, \emph{Spatial analysis along networks: statistical
  and computational methods}.\hskip 1em plus 0.5em minus 0.4em\relax John Wiley
  \& Sons, 2012.

\bibitem{jiang2019hidden}
Z.~Jiang and A.~M. Sainju, ``Hidden markov contour tree: A spatial structured
  model for hydrological applications,'' in \emph{Proceedings of the 25th ACM
  SIGKDD International Conference on Knowledge Discovery \& Data Mining}, 2019,
  pp. 804--813.

\bibitem{jiang2019geographical}
Z.~Jiang, M.~Xie, and A.~M. Sainju, ``Geographical hidden markov tree,''
  \emph{IEEE Transactions on Knowledge and Data Engineering}, 2019.

\bibitem{sainju2020hidden}
A.~M. Sainju, W.~He, and Z.~Jiang, ``A hidden markov contour tree model for
  spatial structured prediction,'' \emph{IEEE Transactions on Knowledge and
  Data Engineering}, 2020.

\bibitem{pan2010survey}
S.~J. Pan and Q.~Yang, ``A survey on transfer learning,'' \emph{IEEE
  Transactions on knowledge and data engineering}, vol.~22, no.~10, pp.
  1345--1359, 2010.

\bibitem{daume2006domain}
H.~Daume~III and D.~Marcu, ``Domain adaptation for statistical classifiers,''
  \emph{Journal of Artificial Intelligence Research}, vol.~26, pp. 101--126,
  2006.

\bibitem{hall1997introduction}
D.~L. Hall and J.~Llinas, ``An introduction to multisensor data fusion,''
  \emph{Proceedings of the IEEE}, vol.~85, no.~1, pp. 6--23, 1997.

\bibitem{waltz2001principle}
E.~Waltz, ``The principle and practice of image and spatial data fusion, ser.
  handbook of multisensor data fusion, d. hall,'' \emph{J. Llinas. CRC Press
  LLC}, vol.~4, 2001.

\bibitem{khaleghi2013multisensor}
B.~Khaleghi, A.~Khamis, F.~O. Karray, and S.~N. Razavi, ``Multisensor data
  fusion: A review of the state-of-the-art,'' \emph{Information Fusion},
  vol.~14, no.~1, pp. 28--44, 2013.

\bibitem{bleiholder2009data}
J.~Bleiholder and F.~Naumann, ``Data fusion,'' \emph{ACM Computing Surveys
  (CSUR)}, vol.~41, no.~1, p.~1, 2009.

\bibitem{dong2009data}
X.~L. Dong and F.~Naumann, ``Data fusion: resolving data conflicts for
  integration,'' \emph{Proceedings of the VLDB Endowment}, vol.~2, no.~2, pp.
  1654--1655, 2009.

\bibitem{karpatne2016monitoring}
A.~Karpatne, Z.~Jiang, R.~R. Vatsavai, S.~Shekhar, and V.~Kumar, ``Monitoring
  land-cover changes: A machine-learning perspective,'' \emph{IEEE Geoscience
  and Remote Sensing Magazine}, vol.~4, no.~2, pp. 8--21, 2016.

\bibitem{goodfellow2016deep}
I.~Goodfellow, Y.~Bengio, and A.~Courville, ``Deep learning (adaptive
  computation and machine learning series),'' \emph{Adaptive Computation and
  Machine Learning series}, p. 800, 2016.

\bibitem{zhang2016deep}
L.~Zhang, L.~Zhang, and B.~Du, ``Deep learning for remote sensing data: A
  technical tutorial on the state of the art,'' \emph{IEEE Geoscience and
  Remote Sensing Magazine}, vol.~4, no.~2, pp. 22--40, 2016.

\bibitem{zhu2017deep}
X.~X. Zhu, D.~Tuia, L.~Mou, G.-S. Xia, L.~Zhang, F.~Xu, and F.~Fraundorfer,
  ``Deep learning in remote sensing: a review,'' \emph{arXiv preprint
  arXiv:1710.03959}, 2017.

\bibitem{ronneberger2015u}
O.~Ronneberger, P.~Fischer, and T.~Brox, ``U-net: Convolutional networks for
  biomedical image segmentation,'' in \emph{International Conference on Medical
  image computing and computer-assisted intervention}.\hskip 1em plus 0.5em
  minus 0.4em\relax Springer, 2015, pp. 234--241.

\bibitem{deng2009imagenet}
J.~Deng, W.~Dong, R.~Socher, L.-J. Li, K.~Li, and L.~Fei-Fei, ``Imagenet: A
  large-scale hierarchical image database,'' in \emph{Computer Vision and
  Pattern Recognition, 2009. CVPR 2009. IEEE Conference on}.\hskip 1em plus
  0.5em minus 0.4em\relax IEEE, 2009, pp. 248--255.

\bibitem{iarpa}
{IARPA}, ``Functional map of the world challenge,''
  \url{https://www.iarpa.gov/challenges/fmow.html}.

\bibitem{yang2010bag}
Y.~Yang and S.~Newsam, ``Bag-of-visual-words and spatial extensions for
  land-use classification,'' in \emph{Proceedings of the 18th SIGSPATIAL
  international conference on advances in geographic information
  systems}.\hskip 1em plus 0.5em minus 0.4em\relax ACM, 2010, pp. 270--279.

\bibitem{karpatne2017theory}
A.~Karpatne, G.~Atluri, J.~H. Faghmous, M.~Steinbach, A.~Banerjee, A.~Ganguly,
  S.~Shekhar, N.~Samatova, and V.~Kumar, ``Theory-guided data science: A new
  paradigm for scientific discovery from data,'' \emph{IEEE Transactions on
  Knowledge and Data Engineering}, vol.~29, no.~10, pp. 2318--2331, 2017.

\bibitem{shekhar2012spatial}
S.~Shekhar, V.~Gunturi, M.~R. Evans, and K.~Yang, ``Spatial big-data challenges
  intersecting mobility and cloud computing,'' in \emph{Proceedings of the
  Eleventh ACM International Workshop on Data Engineering for Wireless and
  Mobile Access}.\hskip 1em plus 0.5em minus 0.4em\relax ACM, 2012, pp. 1--6.

\bibitem{vatsavai2012spatiotemporal}
R.~R. Vatsavai, A.~Ganguly, V.~Chandola, A.~Stefanidis, S.~Klasky, and
  S.~Shekhar, ``Spatiotemporal data mining in the era of big spatial data:
  algorithms and applications,'' in \emph{Proceedings of the 1st ACM SIGSPATIAL
  international workshop on analytics for big geospatial data}.\hskip 1em plus
  0.5em minus 0.4em\relax ACM, 2012, pp. 1--10.

\bibitem{planthaber2012earthdb}
G.~Planthaber, M.~Stonebraker, and J.~Frew, ``Earthdb: scalable analysis of
  modis data using scidb,'' in \emph{Proceedings of the 1st ACM SIGSPATIAL
  International Workshop on Analytics for Big Geospatial Data}.\hskip 1em plus
  0.5em minus 0.4em\relax ACM, 2012, pp. 11--19.

\bibitem{eldawy2015spatialhadoop}
A.~Eldawy and M.~F. Mokbel, ``Spatialhadoop: A mapreduce framework for spatial
  data,'' in \emph{Data Engineering (ICDE), 2015 IEEE 31st International
  Conference on}.\hskip 1em plus 0.5em minus 0.4em\relax IEEE, 2015, pp.
  1352--1363.

\bibitem{yu2015geospark}
J.~Yu, J.~Wu, and M.~Sarwat, ``Geospark: A cluster computing framework for
  processing large-scale spatial data,'' in \emph{Proceedings of the 23rd
  SIGSPATIAL International Conference on Advances in Geographic Information
  Systems}.\hskip 1em plus 0.5em minus 0.4em\relax ACM, 2015, p.~70.

\bibitem{prasad2015vision}
S.~K. Prasad, M.~McDermott, S.~Puri, D.~Shah, D.~Aghajarian, S.~Shekhar, and
  X.~Zhou, ``A vision for gpu-accelerated parallel computation on geo-spatial
  datasets,'' \emph{SIGSPATIAL Special}, vol.~6, no.~3, pp. 19--26, 2015.

\bibitem{puri2013efficient}
S.~Puri and S.~K. Prasad, ``Efficient parallel and distributed algorithms for
  gis polygonal overlay processing,'' in \emph{Parallel and Distributed
  Processing Symposium Workshops \& PhD Forum (IPDPSW), 2013 IEEE 27th
  International}.\hskip 1em plus 0.5em minus 0.4em\relax IEEE, 2013, pp.
  2238--2241.

\bibitem{zhang2012speeding}
J.~Zhang and S.~You, ``Speeding up large-scale point-in-polygon test based
  spatial join on gpus,'' in \emph{Proceedings of the 1st ACM SIGSPATIAL
  International Workshop on Analytics for Big Geospatial Data}.\hskip 1em plus
  0.5em minus 0.4em\relax ACM, 2012, pp. 23--32.

\bibitem{googleearthengine}
{Google Earth Engine Team}, ``Google earth engine: A planetary-scale
  geo-spatial analysis platform,'' \url{https://earthengine.google.com}, 12
  2015.

\bibitem{shekhar2003spatial}
S.~Shekhar and S.~Chawla, \emph{Spatial Databases: A Tour}, ser. An Alan R. Apt
  book.\hskip 1em plus 0.5em minus 0.4em\relax Prentice Hall, 2003.

\end{thebibliography}
%

\end{document}